%% file: main.tex
\documentclass[10pt]{article}

\usepackage[utf8]{inputenc}
\usepackage[T1]{fontenc}
\usepackage{lmodern}
\usepackage[a4paper,hmargin=2.5cm,vmargin=1.9cm]{geometry}
\usepackage{amsmath}
\usepackage{amssymb}
\usepackage{graphicx}
\usepackage{booktabs}
\usepackage{multirow}
\usepackage{array}
\usepackage{makecell}
\usepackage{adjustbox}
\usepackage{float}
\usepackage[strict]{changepage}
\usepackage{xcolor}
\usepackage{tikz}
\usepackage{pgfplots}
\pgfplotsset{compat=1.18}
\usepackage[font=small,labelfont=bf]{caption}
\usepackage[numbers,sort&compress]{natbib}
\usepackage{authblk}
\usepackage{url}
\usepackage[hidelinks]{hyperref}
\usepackage{orcidlink}
\newlength{\extralength}
\newcommand{\keywords}[1]{\par\medskip\noindent\textbf{Keywords:} #1}

\title{Interpretable and Calibrated Classification of Clinical Data Using Supervised Feature Binarization}

\author[1,3]{Antony Garcia\,\orcidlink{0000-0002-5795-2104}}
\author[4,5,6,7]{Adri\'an Noriega de la Colina\,\orcidlink{0000-0001-9933-8164}}
\author[2]{Gabrielle Britton\,\orcidlink{0000-0002-1758-2495}}
\author[1,*]{Xinming Huang\,\orcidlink{0000-0003-0584-3448}}

\affil[1]{Department of Electrical and Computer Engineering, Worcester Polytechnic Institute, Worcester, MA 01609, USA}
\affil[2]{Centro de Vacunaci\'on e Investigaci\'on (CEVAXIN), Panama City, Panama}
\affil[3]{Facultad de Ingenier\'ia El\'ectrica, Universidad Tecnol\'ogica de Panam\'a, Panama City 0819-07289, Panama}
\affil[4]{Massachusetts Institute of Technology, Cambridge, MA 02139, USA}
\affil[5]{Broad Institute of MIT and Harvard, Cambridge, MA 02142, USA}
\affil[6]{McGill University, Montreal, QC H3A 0G4, Canada}
\affil[7]{Montreal Neurological Institute-Hospital, 3801 University Street, Montreal, QC H3A 2B4, Canada}
\affil[*]{Correspondence: \href{mailto:xhuang@wpi.edu}{xhuang@wpi.edu}}

\date{}

\begin{document}

\maketitle

\begin{abstract}
Black-box models limit the adoption of artificial intelligence in medicine because their predictions are difficult to interpret and reproduce. We present a statistically grounded framework for interpretable, rule-based clinical classification using the Bernoulli Na\"ive Bayes (BNB) model. Supervised chi-square-guided binarization converts continuous variables into binary indicators by selecting thresholds that maximize association with the clinical outcome within the training folds, which allows BNB to operate on continuous medical data without sacrificing transparency. On three benchmark datasets, Pima Indians Diabetes, Wisconsin Breast Cancer, and Heart Failure Prediction, the framework reached areas under the receiver operating characteristic curve of 0.800, 0.984, and 0.919, respectively. Probabilistic reliability was assessed with a leakage-safe cross-validated calibration analysis reporting Brier score and calibration intercept and slope, and post-hoc beta calibration improved probability calibration across datasets. These results indicate that an interpretable, statistically motivated framework can perform comparably to more complex models while providing explicit decision rules expressed in clinical units and calibrated risk estimates. A complete worked example further shows that model inference can be reproduced from a printed reference table using only basic arithmetic, without software or proprietary tools, supporting trustworthy and auditable use of artificial intelligence in clinical settings.
\keywords{explainable AI; Bernoulli Na\"ive Bayes; feature binarization; chi-square thresholding; interpretable machine learning; medical classification; threshold optimization; probability calibration}
\end{abstract}

\input{section_01_introduction}

\input{section_02_related_work}

\input{section_03_methods}

\input{section_04_results}
\input{section_05_discussion}
\input{section_06_conclusion_future_work}

\section*{Author Contributions}
Conceptualization, A.G.; methodology, A.G.; software, A.G.; formal analysis, A.G., A.N., G.B. and X.H.; data curation, A.G.; visualization, A.G.; writing---original draft preparation, A.G.; writing---review and editing, A.G., A.N., G.B. and X.H.; supervision, A.N. and X.H. All authors have read and agreed to this version of the manuscript.

\section*{Funding}
This research received no external funding.

\section*{Ethics Statement}
Institutional review and informed consent were not applicable. This study used only publicly available, de-identified benchmark datasets and did not involve new data collection, direct interaction with human participants, or access to identifiable private information.

\section*{Data Availability}
All datasets analyzed in this study are publicly available benchmark datasets: the Pima Indians Diabetes Database \cite{smithUsingADAPLearning1988}, the Breast Cancer Wisconsin (Diagnostic) Database \cite{williamwolbergBreastCancerWisconsin1993}, and the Heart Failure Prediction Dataset \cite{janosiHeartFailurePrediction2021}. No new data were generated. The code developed for this study is available from the corresponding author on reasonable request.

\section*{Conflicts of Interest}
The authors declare no conflicts of interest.

\section*{Declaration of Generative AI and AI-Assisted Technologies}
The authors did not use generative AI or AI-assisted technologies in the preparation of this manuscript.

\section*{Abbreviations}
\noindent
\begin{tabular}{@{}ll}
AI & Artificial intelligence\\
AUC & Area under the curve\\
BMI & Body mass index\\
BNB & Bernoulli Na\"ive Bayes\\
CI & Confidence interval\\
CNB & Categorical Na\"ive Bayes\\
ECG & Electrocardiogram\\
NB & Na\"ive Bayes\\
OOF & Out-of-fold\\
PIDD & Pima Indians Diabetes Database\\
RBF & Radial basis function\\
ROC & Receiver operating characteristic\\
SVM & Support vector machine\\
WHO & World Health Organization
\end{tabular}

\bibliographystyle{unsrtnat}
\bibliography{references}

\end{document}

%% file: section_01_introduction.tex
\section{Introduction}

Supervised classification is a fundamental task in medical data analysis. Its objective is the assignment of patients to clinically meaningful classes on the basis of their observed characteristics \cite{moreno-ibarraClassificationDiseasesUsing2021}. In practice, information such as demographics, medical history, and clinical measurements is combined to determine whether a patient has a condition or is at risk of developing one \cite{anComprehensiveReviewMachine2023}.

These predictions are not merely technical outputs, because diagnosis, treatment planning, and patient outcomes can be directly influenced by them \cite{ennabEnhancingInterpretabilityAccuracy2024}. Several practical challenges are also presented by medical datasets: both continuous and categorical variables are often included, missing values may be present, and skewed class distributions are common \cite{chenDealingMissingImbalanced2023,razzaghiMultilevelWeightedSupport2016}. For a classification model to be useful in clinical settings, these characteristics must be handled while maintaining sufficient interpretability for the resulting decisions to be trusted and understood by clinicians \cite{kumarExplainableMachineLearning2024}.

Artificial intelligence has demonstrated remarkable potential for improving diagnostic accuracy across a wide range of clinical applications \cite{richensImprovingAccuracyMedical2020,fuseArtificialIntelligenceClinical2025,ahsanMachineLearningBasedDiseaseDiagnosis2022}. However, many high-performing models, especially deep learning algorithms, are considered black boxes and may generalize poorly to diverse patient populations \cite{tjoaSurveyExplainableArtificial2021}. This lack of transparency can limit clinician trust and complicate regulatory adoption \cite{topolHighperformanceMedicineConvergence2019,kellyKeyChallengesDelivering2019}, making interpretable probabilistic models attractive for medical decision support.

Among supervised learning approaches, probabilistic classifiers are considered an important family because predictions are based on estimated probability distributions over class labels. This perspective is especially useful in clinical settings, where uncertainty is common. A notable example is the Bayesian network, in which conditional dependencies among features are represented and data analysis and decision-making under uncertainty are supported \cite{perezSupervisedClassificationConditional2006}. Within this broader family, conditional independence among features is assumed by the Naïve Bayes model, resulting in models that are easy to interpret and that often perform well even when the independence assumption is not always satisfied \cite{domingosOptimalitySimpleBayesian1997}.

Several likelihood variants adapted to different types of data are included in the Naïve Bayes family: Bernoulli for binary features, Gaussian for continuous, normally distributed features, Multinomial for count-based data, and Categorical for discrete categorical features. High effectiveness has been reported for Bernoulli models in text classification tasks where the presence or absence of words is indicated by features \cite{mccallum_comparison_1998,singhComparisonMultinomialBernoulli2019,kharismaComparisonNaiveBayes2021}, while Gaussian and Multinomial variants are generally preferred for continuous and frequency-based inputs, respectively \cite{xuBayesianNaiveBayes2018,kibriyaMultinomialNaiveBayes2004}. Categorical models handle discrete features directly without any assumption of ordering or distance between categories, making them well-adapted to tabular data with nominal features.

Despite its simplicity and interpretability, limited application in medical data analysis has been seen for the Bernoulli variant because of its reliance on binary features. Continuous variables, such as age, blood pressure, and glucose levels, are often contained in medical datasets and cannot be directly used with Bernoulli models without preprocessing. As a result, the use of Bernoulli Naïve Bayes models in clinical settings has been restricted, despite the high value placed on interpretability, as preprocessing steps can introduce unnecessary complexity and introduce potential bias.

This limitation is addressed in the present work through the introduction of a statistical binarization method based on chi-square ($\chi^2$) independence tests. Continuous features are transformed by the proposed method into binary indicators through the selection of thresholds at which the association with the outcome is least likely to occur by chance under the independence assumption.

During method development, several thresholding criteria were evaluated, including mutual information, information gain, Gini gain, ROC-based criteria, Otsu thresholding, and distribution-based cut points. Among these alternatives, the $\chi^2$ criterion was found to deliver the most consistent performance across the studied datasets while remaining closely aligned with the independence-testing rationale of the Bernoulli model. Threshold optimization was performed exclusively within the training folds, and final performance was evaluated only on held-out data to avoid information leakage and to obtain an unbiased estimate of generalization.

By this strategy, Bernoulli models can be applied effectively to continuous medical data while their interpretability is preserved. The main contributions of this paper are: (1) a clear formulation of the challenges involved in applying Bernoulli models to medical datasets with continuous features; (2) a $\chi^2$-based binarization approach for determining feature thresholds; (3) an empirical evaluation of the proposed method on three diverse and clinically relevant datasets; (4) binary features that correspond to explicit decision boundaries, thereby supporting transparent interpretation in clinical settings; and (5) a complete patient-level worked example demonstrating that the full classification pathway, from raw feature values to posterior probability, can be reproduced by a clinician using a printed reference table and a calculator, without requiring software access.

Alignment is also shown between this emphasis on transparent threshold-based rules and recent international frameworks, including the EU Artificial Intelligence Act \cite{aboyNavigatingEUAI2024} and the WHO guidance on the ethics and governance of artificial intelligence for health \cite{noauthor_ethics_2021}, in which explainability, fairness, and reproducibility are identified as key requirements for trustworthy clinical AI. By embedding interpretability at the feature-engineering stage instead of relying only on post-hoc explanations, more transparent deployment within diverse clinical settings is intended to be supported by the proposed approach.

The remainder of this paper is organized as follows. Section 2 reviews prior work on Naïve Bayes models, supervised binarization, and interpretable clinical prediction. Section 3 describes the materials and methods, including the datasets, preprocessing procedures, Bernoulli and Categorical Naïve Bayes models, the proposed $\chi^2$-based threshold-selection method, alternative thresholding strategies, and the evaluation protocol. Section 4 presents the comparative results, calibration analysis, and explainability analyses for the three medical datasets. Section 5 discusses the main implications and limitations of the study. Section 6 concludes the paper and outlines directions for future work.

%% file: section_02_related_work.tex
\section{Related Work}

Naïve Bayes classifiers are one of the fundamental techniques of machine learning, often praised for their simplicity, efficiency, and interpretability \cite{domingosOptimalitySimpleBayesian1997,zhang_exploring_2004}. Although Naïve Bayes assumes conditional independence between features, it has consistently achieved good performance in medical classification tasks, especially when paired with effective feature transformations.

In the medical field, Bernoulli Naïve Bayes has proven to be effective for tasks involving binary or thresholded features. BNB has demonstrated competitive performance in clinical classification tasks when preprocessing is appropriately designed \cite{weegarUsingMachineLearning2020,taghizadehBreastCancerPrediction2022,patraPredictionLungCancer2020,aliFeatureDrivenDecisionSupport2019,guptaEarlyAutismSpectrum2025,singhOptimizingEarlyDiagnosis2024}. Naïve Bayes has been successfully applied to predict lung cancer \cite{patraPredictionLungCancer2020}, detect cardiovascular diseases \cite{aliFeatureDrivenDecisionSupport2019}, and autism screening \cite{guptaEarlyAutismSpectrum2025}, showing their relevance in real-world medical contexts where interpretability matters.

One of the main challenges in using BNB for clinical data lies in handling continuous biomarkers, such as glucose and cholesterol levels. Since BNB requires binary inputs, these continuous values must be converted into meaningful binarized features. In contrast, supervised methods use the target variable to identify thresholds that maximize predictive power. Unlike multi-interval discretization methods, the proposed approach focuses on identifying a single outcome-guided cutoff per feature, aligning with the binary structure required by Bernoulli Naïve Bayes \cite{fayyad_multinterval_1993,kerber_chimerge_1992,tayModifiedChi2Algorithm2002,suExtendedChi2Algorithm2005}. This study demonstrates that statistically guided binarization provides a practical way to convert continuous clinical measurements into interpretable decision rules. Prior work suggests that supervised discretization can enhance predictive discrimination while preserving model interpretability \cite{liuDiscretizationEnablingTechnique2002,ramirez-gallegoDataDiscretizationTaxonomy2016,kotsiantis_discretization_2006,garciaSurveyDiscretizationTechniques2013}.

Recent advances have further applied feature binarization to high-dimensional and domain-specific medical datasets. Binarized clinical predictors, for instance, have been used to build interpretable and well-calibrated heart disease risk models \cite{salehiInterpretableHeartDisease2026}, while data-driven binarization thresholds have been shown to improve classification performance on high-dimensional microbiome data \cite{leeSHAPbasedBinarizationEnhances2025}. Hybrid methods combining discretization with evolutionary feature selection on high-dimensional data \cite{tranNewRepresentationPSO2018} or with ensemble classifiers \cite{pekerApplicationChisquareDiscretization2021} have also been reported. Such techniques demonstrate the potential of well-designed binarization techniques to transform raw clinical data into actionable interpretations without a loss of interpretability.

From a clinical modeling perspective, threshold-based transformations yield clear, rule-like representations that can be readily interpreted and audited. Compared with black-box models, statistical binarization in BNB results in understandable, rule-based outputs that can be justified and interpreted by clinicians themselves. This property aligns with commonly stated goals of interpretable machine learning, which emphasize simplicity, accountability, and dependability in clinical decision settings \cite{rudinStopExplainingBlack2019,caruanaIntelligibleModelsHealthCare2015}. Work on explainability \cite{holzingerCausabilityExplainabilityArtificial2019,ghassemiReviewChallengesOpportunities2020,guidottiSurveyMethodsExplaining2019} has frequently discussed the trade-off between predictive performance and interpretability, and several authors argue that inherently interpretable models are preferable for clinical application. Building on this, the combination of $\chi^2$-based statistical binarization with BNB has been proposed in this study as a statistically sound and inherently interpretable solution.

%% file: section_03_methods.tex
\section{Materials and Methods}
\label{sec:Methods}

\subsection{Overview}

This study presents a method for binarizing continuous features, making it possible to use the Bernoulli Naïve Bayes classifier with datasets that include continuous values. The approach is based on statistical hypothesis testing: for each feature, multiple thresholds are assessed using $\chi^2$ tests of independence to measure the association between the binarized feature and the target variable. The threshold with the lowest p-value from the $\chi^2$ test is chosen, as it marks the point where the observed relationship is least consistent with independence.

It should be noted that the p-value in this context does not indicate the probability that a feature is predictive nor the strength of the association. Instead, it measures how much the observed distribution differs from what would be expected if the feature and target variable were independent, and is used as a practical guide for selecting thresholds in supervised learning.

To evaluate the performance of the chi-square-based binarization method, other thresholding strategies were also implemented and compared, including mutual information, information gain, Gini gain, ROC-based criteria, Otsu thresholding, and distribution-based cut points. 

The classification performance of the proposed method was compared with several reference models from the literature across three publicly available clinical datasets. Dataset characteristics and dataset-specific preprocessing protocols are detailed in Section~\ref{subsec:datasets_preprocessing}.

\subsection{Datasets and Dataset-Specific Preprocessing}
\label{subsec:datasets_preprocessing}

Experiments were conducted on three publicly available medical datasets with binary targets: the Pima Indians Diabetes Dataset (PIDD), the Wisconsin Breast Cancer Dataset, and the Heart Failure Prediction Dataset. These datasets are widely used benchmarks in clinical machine learning and were selected to cover settings with continuous features, mixed feature types, and missing-value challenges.

\begin{itemize}
    \item \textbf{Pima Indians Diabetes Dataset (PIDD)}: contains 768 records with 8 clinical predictors and a binary diabetes outcome \cite{smithUsingADAPLearning1988}.
    \item \textbf{Wisconsin Breast Cancer Dataset}: contains 569 records with 30 predictors derived from digitized fine-needle aspirate images and a binary benign/malignant label \cite{williamwolbergBreastCancerWisconsin1993}.
    \item \textbf{Heart Failure Prediction Dataset}: contains 918 records with 11 predictors, including both numerical and categorical clinical variables, and a binary heart-disease outcome \cite{janosiHeartFailurePrediction2021}.
\end{itemize}

Summary statistics and class distributions for the three datasets are reported in Tables \ref{tab:diabetes_risk_factors}, \ref{tab:breast_cancer}, and \ref{tab:heart_dataset}.

For PIDD, zeros in Glucose, Blood Pressure, Skin Thickness, Insulin, and BMI were treated as missing values following prior studies. Two evaluation settings were considered for BNB: (i) mean-imputation preprocessing for direct comparability with published baselines, and (ii) a missing-aware BNB likelihood that excludes missing features from the product term at prediction time. This second setting preserves incomplete records and was tested specifically for this dataset.

For the Heart Failure dataset, which contains both numerical and categorical predictors, Categorical Naïve Bayes (CNB) was used for the categorical predictors using the encoding described in Table~\ref{tab:heart_dataset_encoding_summary}, while continuous predictors were evaluated in the BNB thresholding framework. As will be shown in the next section, CNB and BNB are mathematically equivalent for binary features, so the same $\chi^2$-based thresholding method can be applied to the continuous features, while categorical features are retained in their original form for CNB evaluation.

\input{table_1}

\subsection{Bernoulli Naïve Bayes Classifier}

The BNB classifier is a probabilistic model specifically designed for feature vectors with binary values. It belongs to the broader family of Naïve Bayes classifiers, which rely on the assumption of conditional independence among features given the class label. Unlike Gaussian Naïve Bayes, which models continuous features using normal distributions, BNB models each binary feature using a Bernoulli distribution.

\input{table_2}

\input{table_3}

Let \( \mathbf{x} = (x_1, x_2, \dots, x_n) \in \{0,1\}^n \) denote a binary feature vector, and let \( \{C_1, \dots, C_K\} \) represent the set of possible class labels. For each feature \( x_i \), the class-conditional likelihood under class \( C_k \) is modeled as:

\begin{equation}
\mathbb{P}(x_i \mid C_k) = p_{k,i}^{x_i} (1 - p_{k,i})^{1 - x_i},
\end{equation}

where \( p_{k,i} = \mathbb{P}(x_i = 1 \mid C_k) \) denotes the probability that feature \( x_i \) is active (equals 1) given class \( C_k \).

Assuming feature independence, the joint likelihood of the full feature vector \( \mathbf{x} \) conditioned on class \( C_k \) becomes:

\begin{equation}
\mathbb{P}(\mathbf{x} \mid C_k) = \prod_{i=1}^{n} p_{k,i}^{x_i} (1 - p_{k,i})^{1 - x_i}.
\end{equation}

The posterior probability \( \mathbb{P}(C_k \mid \mathbf{x}) \) of class \( C_k \) given the observed feature vector \( \mathbf{x} \) is obtained via Bayes' theorem. The predicted class label \( \hat{y} \) is the one that maximizes this posterior:

\begin{equation}
\hat{y} = \arg\max_{k \in \{1, \dots, K\}} \mathbb{P}(C_k) \prod_{i=1}^{n} p_{k,i}^{x_i} (1 - p_{k,i})^{1 - x_i}.
\end{equation}

Since BNB requires all features to be binary, applying it to continuous-valued datasets requires a binarization step. The method used for this transformation is detailed in the following subsection.

In this study, a modified version of the BNB model was employed in the Pima Indians Diabetes Dataset, which contains missing values in several features. To address this, the likelihood computation was adjusted to exclude missing values from the product term. Let \( \mathcal{I}_{\text{obs}} \subseteq \{1, \dots, n\} \) denote the indices of the observed (non-missing) features for a given instance. The modified likelihood is then computed as:

\begin{equation}
\mathbb{P}(\mathbf{x}_{\text{obs}} \mid C_k) = \prod_{i \in \mathcal{I}_{\text{obs}}} p_{k,i}^{x_i} (1 - p_{k,i})^{1 - x_i}.
\end{equation}

This adjustment effectively marginalizes over the missing components, preventing them from influencing the posterior estimation and thereby maintaining the integrity of the prediction. The results of this evaluation can be seen in Table~\ref{tab:diabetes_statistical_comparison}.

\subsection{Categorical Naïve Bayes Classifier}

The Categorical Naïve Bayes classifier is another variant of the Naïve Bayes family, designed for discrete-valued features with more than two categories. While BNB is limited to binary features, CNB extends the likelihood model to handle features with multiple discrete values.

Let \( \mathbf{x} = (x_1, x_2, \dots, x_n) \) be a feature vector where each feature \( x_i \in \{v_1^{(i)}, v_2^{(i)}, \dots, v_{r_i}^{(i)}\} \) takes values from a finite set of \( r_i \) discrete categories. Given a set of classes \( \{C_1, \dots, C_K\} \), the likelihood of a specific categorical value \( x_i = v_j^{(i)} \) under class \( C_k \) is defined as:

\begin{equation}
\mathbb{P}(x_i = v_j^{(i)} \mid C_k) = \theta_{k,ij},
\end{equation}

where \( \theta_{k,ij} \) denotes the class-conditional probability of observing value \( v_j^{(i)} \) for feature \( x_i \) in class \( C_k \), and satisfies the constraint:

\begin{equation}
\sum_{j=1}^{r_i} \theta_{k,ij} = 1 \quad \text{for all } i \text{ and } k.
\end{equation}

Assuming conditional independence of features given the class label, the joint likelihood of the feature vector \( \mathbf{x} \) is:

\begin{equation}
\mathbb{P}(\mathbf{x} \mid C_k) = \prod_{i=1}^{n} \theta_{k,ij} \quad \text{where } x_i = v_j^{(i)}.
\end{equation}

The posterior probability \( \mathbb{P}(C_k \mid \mathbf{x}) \) is proportional to the product of the prior and the likelihood, as computed below:

\begin{equation}
\mathbb{P}(C_k \mid \mathbf{x}) \propto \mathbb{P}(C_k) \prod_{i=1}^{n} \theta_{k,ij}.
\end{equation}

The predicted class label \( \hat{y} \) is determined by:

\begin{equation}
\hat{y} = \arg\max_{k \in \{1, \dots, K\}} \mathbb{P}(C_k) \prod_{i=1}^{n} \theta_{k,ij}.
\end{equation}

If CNB is applied to binary features, it yields results equivalent to those of the BNB classifier, as both models reduce to the same probabilistic formulation under such conditions. This model was used to evaluate the Heart Failure Prediction Dataset, where categorical features were retained in their original form. Table~\ref{tab:heart_dataset_encoding_summary} summarizes the encoding scheme applied to categorical features prior to CNB evaluation.

\subsubsection{Equivalence of Categorical and Bernoulli Naïve Bayes for Binary Features}

Assume that each feature is binary, \(x_i \in \{0, 1\}\), with the two possible values defined as \(v_1^{(i)} = 1\) (``active'') and \(v_2^{(i)} = 0\) (``inactive''). For Categorical Naïve Bayes the class-conditional likelihood of a feature vector \(\mathbf{x}\) under class \(C_k\) is

\[
\mathbb{P}(\mathbf{x}\mid C_k)
      =\prod_{i=1}^{n}\theta_{k,i1}^{\,\mathbf{I}[x_i=1]}\,
                        \theta_{k,i2}^{\,\mathbf{I}[x_i=0]},
\]

where the parameters satisfy \(\theta_{k,i1} + \theta_{k,i2} = 1\).  
Setting  

\[
p_{k,i} \;=\; \theta_{k,i1}, 
\qquad
1-p_{k,i} \;=\; \theta_{k,i2},
\]

and noting that \(\mathbf{I}[x_i=1] = x_i\) and \(\mathbf{I}[x_i=0] = 1 - x_i\), the likelihood becomes  

\[
\mathbb{P}(\mathbf{x}\mid C_k)
      =\prod_{i=1}^{n}
         p_{k,i}^{\,x_i}\,
         (1-p_{k,i})^{\,1-x_i},
\]

which is precisely the Bernoulli Naïve Bayes formulation.
Because both models share the same class priors and employ an identical maximum-a-posteriori decision rule, they yield the same posterior probabilities and therefore identical predictions.

This result justifies treating Categorical NB as a seamless extension of Bernoulli NB to datasets that contain multi-level categorical variables while preserving interpretability for binary features.

\subsection{Optimal Threshold Selection for Feature Binarization}

Consider a dataset represented as:

\[
\mathcal{D} = (x, y)
\]

where \( x \) is a matrix of continuous features, and \( y \) is a binary target vector. The feature matrix is defined as:

\[
x =
\begin{bmatrix}
x_{1,1} & x_{1,2} & \dots & x_{1,m} \\
x_{2,1} & x_{2,2} & \dots & x_{2,m} \\
\vdots & \vdots & \ddots & \vdots \\
x_{n,1} & x_{n,2} & \dots & x_{n,m}
\end{bmatrix}
\in \mathbb{R}^{n \times m}
\]

where \( n \) is the number of samples and \( m \) the number of features. The corresponding binary target vector is given by:

\[
y = [y_1, y_2, \dots, y_n], \quad y_i \in \{0,1\}
\]

Each feature \( x_j \in \mathbb{R}^n \) is processed independently to identify its optimal binarization threshold.

To ensure meaningful comparisons, each feature vector \( x_j \) is first filtered to remove duplicate values. The resulting set of unique values defines the set of threshold candidates:

\[
T_j = \{ x_{i,j} \mid i \in \{1, 2, \dots, n\} \}
\]

For every threshold candidate \( t \in T_j \), a binarized vector \( x_{t,j} \in \{0,1\}^n \) is computed as:

\[
x_{i,j}^{t} =
\begin{cases}
1, & x_{i,j} > t \\
0, & x_{i,j} \leq t
\end{cases}
\quad \forall i \in \{1, \dots, n\}
\]

The relationship between the binarized feature vector \( x_{t,j} \) and the binary target \( y \) is assessed via a chi-square test of independence. A contingency table is constructed for each threshold \( t \):

\[
C =
\begin{bmatrix}
n_{00} & n_{01} \\
n_{10} & n_{11}
\end{bmatrix}
\]

\begin{itemize}
    \item \( n_{00} \): number of samples with \( x_{t,j} = 0 \) and \( y = 0 \)
    \item \( n_{01} \): number of samples with \( x_{t,j} = 0 \) and \( y = 1 \)
    \item \( n_{10} \): number of samples with \( x_{t,j} = 1 \) and \( y = 0 \)
    \item \( n_{11} \): number of samples with \( x_{t,j} = 1 \) and \( y = 1 \)
\end{itemize}

The chi-square statistic is computed as:

\[
\chi^2 = \sum_{i=1}^2 \sum_{j=1}^2 \frac{(O_{ij} - E_{ij})^2}{E_{ij}}
\]

where \( O_{ij} \) are the observed counts and \( E_{ij} \) the expected counts under the null hypothesis of independence (\(H_0\)). For a \(2 \times 2\) table, the degrees of freedom are:

\[
df = (2 - 1)(2 - 1) = 1
\]

The p-value for threshold \( t \), denoted \( p_t \), is then computed as:

\[
p_t = \mathbb{P}(\chi^2 \geq \chi^2_{\text{obs}} \mid H_0) = 1 - F_{\chi^2}(\chi^2_{\text{obs}}, 1)
\]

where \( F_{\chi^2} \) denotes the cumulative distribution function of the chi-square distribution with 1 degree of freedom, and \( \chi^2_{\text{obs}} \) is the observed value of the chi-square statistic calculated from the contingency table for threshold \( t \).

The optimal threshold is selected as the one that minimizes the p-value:

\[
t_j^* = \arg\min_{t \in T_j} p_t
\]

Using this optimal threshold, the final binarized feature vector is computed as:

\[
x_{i,j}^{t_j^*} =
\begin{cases}
1, & x_{i,j} > t_j^* \\
0, & x_{i,j} \leq t_j^*
\end{cases}
\]

The process is repeated for each feature \( j = 1, \dots, m \), yielding an optimal threshold \( t_j^* \) and a corresponding binary feature vector. The final binarized feature matrix is:

\[
x^* =
\begin{bmatrix}
x_{1,1}^{t_1^*} & x_{1,2}^{t_2^*} & \dots & x_{1,m}^{t_m^*} \\
x_{2,1}^{t_1^*} & x_{2,2}^{t_2^*} & \dots & x_{2,m}^{t_m^*} \\
\vdots & \vdots & \ddots & \vdots \\
x_{n,1}^{t_1^*} & x_{n,2}^{t_2^*} & \dots & x_{n,m}^{t_m^*}
\end{bmatrix}
\]

This matrix \( x^* \in \{0,1\}^{n \times m} \) serves as input for training a Bernoulli Naïve Bayes classifier.

\subsection{Other Thresholding Strategies}

To evaluate the robustness of the proposed $\chi^2$-guided thresholding, nine additional well-known binarization strategies were implemented and tested under the same cross-validation procedure. Methods were grouped into two families based on how the threshold is determined.

\textbf{Score-based methods} scan all unique feature values as threshold candidates and select the value that maximises a scoring function applied after binarization:

\begin{itemize}
    \item \textbf{Chi-square ($\chi^2$)}: selects the threshold that minimises the p-value of the $\chi^2$ test of independence between the binarized feature and the binary target.
    \item \textbf{Mutual information}: selects the threshold that maximises $I(X_t; Y)$, the mutual information between the thresholded feature and the target.
    \item \textbf{Gini gain}: selects the threshold that maximises the reduction in Gini impurity induced by the binary split.
    \item \textbf{Information gain}: selects the threshold that maximises the reduction in Shannon entropy (information gain) induced by the split.
    \item \textbf{AUC split}: selects the threshold that maximises $\max(\mathrm{AUC}, 1 - \mathrm{AUC})$, where AUC is computed using the thresholded feature as a univariate binary classifier.
    \item \textbf{Youden's $J$}: selects the threshold that maximises $J = \mathrm{sensitivity} + \mathrm{specificity} - 1$.
\end{itemize}

\textbf{Direct estimators} compute the threshold analytically from the feature distribution without scanning candidates:
\begin{itemize}
    \item \textbf{Otsu threshold}: maximises the between-class variance of the feature histogram, computed via Otsu's method.
    \item \textbf{Median threshold}: sets the threshold at the feature median.
    \item \textbf{Mean threshold}: sets the threshold at the feature mean.
    \item \textbf{K-means midpoint}: fits a two-cluster K-means model ($k = 2$) and sets the threshold at the midpoint between the two cluster centroids.
\end{itemize}

For all methods, thresholding is applied as $x_{i,j}^{t} = \mathbf{I}[x_{i,j} > t]$, and thresholds are estimated exclusively on the training fold within each cross-validation iteration to prevent data leakage.

\subsection{Evaluation Metrics}
\label{subsec:evaluation_metrics}
Classifier performance is evaluated using standard metrics, including accuracy, weighted precision, weighted recall, weighted F1-score, and ROC-AUC \cite{sokolovaSystematicAnalysisPerformance2009}. Special attention is given to the area under the Receiver Operating Characteristic curve, which serves as an overall measure of binary classifier performance. ROC-AUC indicates the model's ability to separate positive and negative classes by quantifying the total area under the ROC curve. As discussed in the literature \cite{fawcettIntroductionROCAnalysis2006}, ROC-AUC is considered the most informative single metric for model evaluation in this study, as it does not depend on the classification threshold or class imbalance.

For each model, predictions from all held-out folds are concatenated to form pooled out-of-fold (OOF) vectors \((y_{\text{true}}, \hat{y}, \hat{p})\). Point estimates are then computed on these pooled OOF predictions. Alongside each point estimate, a 95\% confidence interval (CI) is reported in the result tables using nonparametric bootstrap resampling with replacement (\(B=1000\) replicates). The CI bounds are the percentile limits:
\[
\mathrm{CI}_{95\%} = \left[q_{0.025}\!\left(\hat{\theta}^{*(1:B)}\right),\; q_{0.975}\!\left(\hat{\theta}^{*(1:B)}\right)\right],
\]
where \(\hat{\theta}\) denotes the metric of interest (accuracy, weighted precision, weighted recall, weighted F1-score, or ROC-AUC), and \(\hat{\theta}^{*(b)}\) is the value from bootstrap replicate \(b\).

To compare models statistically, paired tests are performed on matched OOF predictions obtained from the same samples. Each comparison is made against the proposed Bernoulli Naïve Bayes method. All hypothesis tests are two-sided with significance level \(\alpha = 0.05\).

Differences in discrimination between BNB and each comparator were assessed using DeLong's test for correlated ROC-AUC values \cite{delongComparingAreasTwo1988}.
\[
H_0: \mathrm{AUC}_A - \mathrm{AUC}_B = 0
\]
\[
H_1: \mathrm{AUC}_A - \mathrm{AUC}_B \neq 0.
\]
Under \(H_0\), both models have the same discrimination performance. The test statistic is computed as
\[
z = \frac{\widehat{\mathrm{AUC}}_A - \widehat{\mathrm{AUC}}_B}{\mathrm{SE}\!\left(\widehat{\mathrm{AUC}}_A - \widehat{\mathrm{AUC}}_B\right)},
\]
with p-value \(p = 2\Phi(-|z|)\), where \(\Phi\) is the standard normal cumulative distribution function.

Differences in paired classification outcomes were evaluated using McNemar's test \cite{mcnemarNoteSamplingError1947}, based on discordant counts \(b\) and \(c\) computed from matched predictions of BNB and each comparator on the same cases.
\[
H_0: b = c
\]
\[
H_1: b \neq c.
\]
The null hypothesis states that both models have equal error rates on discordant pairs. When \(b+c \ge 25\), the continuity-corrected chi-square form is used:
\[
\chi^2 = \frac{(|b-c|-1)^2}{b+c}.
\]
When \(b+c < 25\), the exact binomial version of McNemar's test is used.

Because relatively few published comparison studies report confidence intervals and the fold-level outputs required for paired statistical testing, additional self-implementations were developed for each comparator model. These implementations followed the methodological instructions described in each paper and were validated by reproducing the reported ROC-AUC as closely as possible before running the statistical comparisons. This can be seen in the tables \ref{tab:diabetes_statistical_comparison}, \ref{tab:cancer_statistical_comparison}, and \ref{tab:heart_statistical_comparison} in the Results section.

\subsection{Calibration Metrics}
\label{subsec:calibration_metrics}

In clinical practice, a predicted probability is not merely a model output; it is a number that a clinician may use to decide whether to refer a patient, initiate treatment, or counsel a family. A model that assigns \(70\%\) risk to a patient should, in practice, identify individuals who experience the outcome roughly seven times out of ten; if it does not, the probability estimate is misleading regardless of how well the model ranks patients relative to one another. In addition to ROC-AUC, calibration was evaluated to assess the reliability of predicted probabilities. While ROC-AUC measures discrimination, that is, the ability to rank cases correctly, it does not assess whether predicted probabilities correspond to observed event frequencies. For clinical decision support, this distinction is important because risk estimates may be used directly for interpretation and threshold-based decisions \cite{vancalsterCalibrationAchillesHeel2019}.

For this study, calibration analysis was incorporated into the same 10-fold cross-validation framework used for model evaluation. For each fold, $\chi^2$-guided thresholds were estimated on the training split only, the corresponding Naïve Bayes model was trained on the transformed training data, and predicted probabilities were generated for the held-out fold. Pooled out-of-fold predictions were then used for calibration assessment, preserving separation between training and testing data and avoiding information leakage.

Let $y_i \in \{0,1\}$ denote the observed outcome for sample $i$, and let $\hat{p}_i \in (0,1)$ denote the predicted probability of the positive class. Using pooled OOF predictions, the following calibration metrics were computed.

\subsubsection{Calibration-in-the-large}

Calibration-in-the-large evaluates whether predicted risks are too high or too low on average \cite{vancalsterCalibrationHierarchyRisk2016}. A simple descriptive summary compares the observed event rate
\[
\bar{y} = \frac{1}{n}\sum_{i=1}^n y_i
\]
with the mean predicted risk
\[
\bar{p} = \frac{1}{n}\sum_{i=1}^n \hat{p}_i.
\]
Good calibration-in-the-large is indicated when $\bar{p} \approx \bar{y}$.

A model-based estimate of calibration-in-the-large is the calibration intercept, obtained from logistic recalibration. An intercept of $0$ is ideal.

\subsubsection{Brier score}
The Brier score measures the mean squared error of probabilistic predictions \cite{huangTutorialCalibrationMeasurements2020}:
\[
\mathrm{Brier} = \frac{1}{n}\sum_{i=1}^n \left(\hat{p}_i - y_i\right)^2.
\]
Lower values indicate better probabilistic accuracy, reflecting both discrimination and calibration.

\subsubsection{Calibration intercept and slope}
Calibration intercept and slope were estimated by fitting a logistic recalibration model to pooled OOF predictions \cite{huangTutorialCalibrationMeasurements2020}:
\[
\mathrm{logit}\!\left(\Pr(y_i=1)\right) = \alpha + \beta \,\mathrm{logit}(\hat{p}_i),
\]
where
\[
\mathrm{logit}(p) = \log\!\left(\frac{p}{1-p}\right).
\]
Here, $\alpha$ is the calibration intercept and $\beta$ is the calibration slope. Ideal calibration corresponds to
\[
\alpha = 0, \qquad \beta = 1.
\]
An intercept $\alpha < 0$ suggests overall overprediction of risk, whereas $\alpha > 0$ suggests underprediction. A slope $\beta < 1$ is commonly associated with overconfident predictions, while $\beta > 1$ indicates underconfident predictions.

\subsubsection{Post-hoc beta calibration}
Because Naïve Bayes models can produce overconfident probabilities, especially when the conditional independence assumption is violated, an additional post-hoc beta calibration step was applied \cite{huangTutorialCalibrationMeasurements2020}. Beta calibration is a parametric probability transformation that generalizes Platt scaling and is well suited to correcting asymmetric distortions in predicted probabilities \cite{kullBetaCalibrationWellfounded2017}.

For each cross-validation fold, beta calibration was fit using training-fold predictions only and then applied to the corresponding held-out fold predictions. In one common parameterization, the calibrated probability $\tilde{p}$ is obtained from the raw probability $\hat{p}$ via
\[
\mathrm{logit}(\tilde{p}) = a \,\log(\hat{p}) + b \,\log(1-\hat{p}) + c,
\]
followed by
\[
\tilde{p} = \frac{1}{1+\exp\!\left[-\left(a \log(\hat{p}) + b \log(1-\hat{p}) + c\right)\right]}.
\]
This transformation was learned on the training fold and applied only to the held-out fold, producing calibrated OOF probabilities that were then evaluated using the same calibration metrics: calibration-in-the-large, Brier score, intercept, and slope.

\subsection{Cross-Validation Strategy}
To ensure a reliable evaluation of the Bernoulli Naïve Bayes classifier, stratified 10-fold cross-validation (\( k=10 \)) is used. In stratified cross-validation, each fold preserves the overall class distribution of the dataset, which is especially important when dealing with imbalanced classes. The dataset is split into ten equally sized folds, maintaining the same proportion of class labels in each. In every iteration, nine folds form the training partition and the remaining fold serves as a held-out validation fold, so that each fold is used once for performance evaluation.

To avoid data leakage, threshold optimization and model fitting are performed exclusively within the training partition in each iteration. Binarization thresholds are calculated only on the training folds and are then applied to both the training data and the held-out validation fold, ensuring that the validation data does not influence preprocessing or parameter estimation.

After binarization, the classifier is trained on the binarized training data and used to predict the corresponding held-out validation fold. Rather than computing each metric separately within every fold and averaging the ten resulting values, the predictions from all held-out folds are concatenated into the pooled out-of-fold vectors defined in Section~\ref{subsec:evaluation_metrics}, and every reported metric is computed once on those pooled vectors. Pooling is required by the evaluation protocol: the paired DeLong and McNemar comparisons operate on matched per-sample predictions rather than on per-fold summaries, and the calibration analysis in Section~\ref{subsec:calibration_metrics} is performed on the same pooled predictions. Because each sample is held out exactly once, the pooled vectors contain one prediction per sample and therefore provide a single leakage-free estimate over the full dataset. For each metric, a 95\% confidence interval is computed via nonparametric bootstrap resampling with replacement (\(B = 1000\) replicates) on these pooled out-of-fold predictions.

\subsection{Comparison study selection criteria}
Studies included for comparison in Tables \ref{tab:diabetes_statistical_comparison}, \ref{tab:cancer_statistical_comparison}, and \ref{tab:heart_statistical_comparison} were identified by searching for research that used the same datasets and reported ROC-AUC values with a k-fold cross-validation approach. Preference was given to studies that described their preprocessing steps, allowing for fair and direct comparison. The selection was based on the relevance of the studies and the availability of results that could be matched to the evaluation criteria of the proposed method.

%% file: table_1.tex
\renewcommand{\arraystretch}{1.5}
\begin{table}[tbp]
\caption{This table provides descriptive statistics for the features of the Pima Indian Diabetes Dataset. The summary includes the range (minimum and maximum values), mean, and standard deviation (Std Dev) for each risk factor across the entire dataset, as well as within two subpopulations: individuals with diabetes and individuals without diabetes.}
\centering

\begin{adjustbox}{max width=\textwidth}

\begin{tabular}{lcccccccccc}
\hline
\multicolumn{1}{c}{\multirow{3}{*}{\textbf{Risk Factor}}} & \multicolumn{1}{l}{\multirow{3}{*}{\textbf{Unit}}} & \multicolumn{3}{c}{\multirow{2}{*}{\textbf{All Population (\textit{N = 768})}}}                  & \multicolumn{6}{c}{\textbf{Diabetes Status}}                                                                                                                  \\ \cmidrule(lr){6-11} 
\multicolumn{1}{c}{}                                      & \multicolumn{1}{l}{}                                    & \multicolumn{3}{c}{}                                                          & \multicolumn{3}{c}{\textbf{Diabetes (\textit{N = 268})}}                                         & \multicolumn{3}{c}{\textbf{No Diabetes (\textit{N = 500})}}                                      \\ \cmidrule(lr){3-11} 
\multicolumn{1}{c}{}                                      & \multicolumn{1}{l}{}                                    & \textbf{Range} & \textbf{Mean} & \textbf{Std Dev} & \textbf{Range} & \textbf{Mean} & \textbf{Std Dev} & \textbf{Range} & \textbf{Mean} & \textbf{Std Dev} \\ \hline
\textbf{Age} & Years & 21.0 - 81.0 & 33.2 & 11.8 & 21.0 - 70.0 & 37.1 & 11.0 & 21.0 - 81.0 & 31.2 & 11.7 \\
\textbf{Pregnancies} & Count & 0.000 - 17.0 & 3.80 & 3.40 & 0.000 - 17.0 & 4.90 & 3.70 & 0.000 - 13.0 & 3.30 & 3.00 \\
\textbf{Glucose} & mg/dL & 0.000 - 199.0 & 120.9 & 32.0 & 0.000 - 199.0 & 141.3 & 31.9 & 0.000 - 197.0 & 110.0 & 26.1 \\
\textbf{Blood Pressure} & mmHg & 0.000 - 122.0 & 69.1 & 19.4 & 0.000 - 114.0 & 70.8 & 21.5 & 0.000 - 122.0 & 68.2 & 18.1 \\
\textbf{Skin Thickness} & mm & 0.000 - 99.0 & 20.5 & 16.0 & 0.000 - 99.0 & 22.2 & 17.7 & 0.000 - 60.0 & 19.7 & 14.9 \\
\textbf{Insulin} & $\mu U$/mL & 0.000 - 846.0 & 79.8 & 115.2 & 0.000 - 846.0 & 100.3 & 138.7 & 0.000 - 744.0 & 68.8 & 98.9 \\
\textbf{BMI} & kg/m\textsuperscript{2} & 0.000 - 67.1 & 32.0 & 7.90 & 0.000 - 67.1 & 35.1 & 7.30 & 0.000 - 57.3 & 30.3 & 7.70 \\
\textbf{Diabetes Pedigree Function} &  & 0.100 - 2.40 & 0.500 & 0.300 & 0.100 - 2.40 & 0.600 & 0.400 & 0.100 - 2.30 & 0.400 & 0.300 \\ \hline
\end{tabular}

\end{adjustbox}
\label{tab:diabetes_risk_factors}
\end{table}

%% file: table_2.tex
\renewcommand{\arraystretch}{1.5}
\begin{table}[tbp]
\centering

\caption{This table provides descriptive statistics for the features of the Wisconsin Breast Cancer Dataset. The summary includes the range (minimum and maximum values), mean, and standard deviation (Std Dev) for each feature across the entire dataset, as well as within two subpopulations: malignant and benign diagnoses.}
\label{tab:breast_cancer}
\begin{adjustbox}{max width=\textwidth}
\begin{tabular}{lcccccccccc}
\hline
\multicolumn{1}{c}{\multirow{3}{*}{\textbf{Feature}}} & \multicolumn{1}{l}{\multirow{3}{*}{\textbf{Units}}} & \multicolumn{3}{c}{\multirow{2}{*}{\textbf{All Population (\textit{N = 569})}}}                  & \multicolumn{6}{c}{\textbf{Diagnosis}}                                                                                           \\ \cmidrule(lr){6-11} 
\multicolumn{1}{c}{}                                  & \multicolumn{1}{l}{}                                & \multicolumn{3}{c}{}                                                          & \multicolumn{3}{c}{\textbf{Malignant (\textit{N = 212})}}                                         & \multicolumn{3}{c}{\textbf{Benign (\textit{N = 357})}}                                             \\ \cmidrule(lr){3-11} 
\multicolumn{1}{c}{}                                  & \multicolumn{1}{l}{}                                & \multicolumn{1}{c}{Range} & \multicolumn{1}{c}{Mean} & \multicolumn{1}{c}{Std dev} & \multicolumn{1}{c}{Range} & \multicolumn{1}{c}{Mean} & \multicolumn{1}{c}{Std dev} & \multicolumn{1}{c}{Range} & \multicolumn{1}{c}{Mean} & \multicolumn{1}{c}{Std dev} \\ 
\hline
\textbf{Radius (Mean)} & mm & 6.98 - 28.1 & 14.1 & 3.52 & 10.9 - 28.1 & 17.5 & 3.20 & 6.98 - 17.9 & 12.1 & 1.78 \\
\textbf{Texture (Mean)} & mm & 9.71 - 39.3 & 19.3 & 4.30 & 10.4 - 39.3 & 21.6 & 3.78 & 9.71 - 33.8 & 17.9 & 4.00 \\
\textbf{Perimeter (Mean)} & mm & 43.8 - 188.5 & 92.0 & 24.3 & 71.9 - 188.5 & 115.4 & 21.9 & 43.8 - 114.6 & 78.1 & 11.8 \\
\textbf{Area (Mean)} & mm$^2$ & 143.5 - 2501.0 & 654.9 & 351.9 & 361.6 - 2501.0 & 978.4 & 367.9 & 143.5 - 992.1 & 462.8 & 134.3 \\
\textbf{Smoothness (Mean)} & ratio & 0.050 - 0.160 & 0.100 & 0.010 & 0.070 - 0.140 & 0.100 & 0.010 & 0.050 - 0.160 & 0.090 & 0.010 \\
\textbf{Compactness (Mean)} & ratio & 0.020 - 0.350 & 0.100 & 0.050 & 0.050 - 0.350 & 0.150 & 0.050 & 0.020 - 0.220 & 0.080 & 0.030 \\
\textbf{Concavity (Mean)} & ratio & 0.000 - 0.430 & 0.090 & 0.080 & 0.020 - 0.430 & 0.160 & 0.080 & 0.000 - 0.410 & 0.050 & 0.040 \\
\textbf{Concave Points (Mean)} & ratio & 0.000 - 0.200 & 0.050 & 0.040 & 0.020 - 0.200 & 0.090 & 0.030 & 0.000 - 0.090 & 0.030 & 0.020 \\
\textbf{Symmetry (Mean)} & ratio & 0.110 - 0.300 & 0.180 & 0.030 & 0.130 - 0.300 & 0.190 & 0.030 & 0.110 - 0.270 & 0.170 & 0.020 \\
\textbf{Fractal Dimension (Mean)} &  & 0.050 - 0.100 & 0.060 & 0.010 & 0.050 - 0.100 & 0.060 & 0.010 & 0.050 - 0.100 & 0.060 & 0.010 \\ \hline
\textbf{Radius (SE)} & mm & 0.110 - 2.87 & 0.410 & 0.280 & 0.190 - 2.87 & 0.610 & 0.350 & 0.110 - 0.880 & 0.280 & 0.110 \\
\textbf{Texture (SE)} & mm & 0.360 - 4.88 & 1.22 & 0.550 & 0.360 - 3.57 & 1.21 & 0.480 & 0.360 - 4.88 & 1.22 & 0.590 \\
\textbf{Perimeter (SE)} & mm & 0.760 - 22.0 & 2.87 & 2.02 & 1.33 - 22.0 & 4.32 & 2.57 & 0.760 - 5.12 & 2.00 & 0.770 \\
\textbf{Area (SE)} & mm$^2$ & 6.80 - 542.2 & 40.3 & 45.5 & 14.0 - 542.2 & 72.7 & 61.4 & 6.80 - 77.1 & 21.1 & 8.84 \\
\textbf{Smoothness (SE)} & ratio & 0.000 - 0.030 & 0.010 & 0.000 & 0.000 - 0.030 & 0.010 & 0.000 & 0.000 - 0.020 & 0.010 & 0.000 \\
\textbf{Compactness (SE)} & ratio & 0.000 - 0.140 & 0.030 & 0.020 & 0.010 - 0.140 & 0.030 & 0.020 & 0.000 - 0.110 & 0.020 & 0.020 \\
\textbf{Concavity (SE)} & ratio & 0.000 - 0.400 & 0.030 & 0.030 & 0.010 - 0.140 & 0.040 & 0.020 & 0.000 - 0.400 & 0.030 & 0.030 \\
\textbf{Concave Points (SE)} & ratio & 0.000 - 0.050 & 0.010 & 0.010 & 0.010 - 0.040 & 0.020 & 0.010 & 0.000 - 0.050 & 0.010 & 0.010 \\
\textbf{Symmetry (SE)} & ratio & 0.010 - 0.080 & 0.020 & 0.010 & 0.010 - 0.080 & 0.020 & 0.010 & 0.010 - 0.060 & 0.020 & 0.010 \\
\textbf{Fractal Dimension (SE)} &  & 0.000 - 0.030 & 0.000 & 0.000 & 0.000 - 0.010 & 0.000 & 0.000 & 0.000 - 0.030 & 0.000 & 0.000 \\ \hline
\textbf{Radius (Worst)} & mm & 7.93 - 36.0 & 16.3 & 4.83 & 12.8 - 36.0 & 21.1 & 4.28 & 7.93 - 19.8 & 13.4 & 1.98 \\
\textbf{Texture (Worst)} & mm & 12.0 - 49.5 & 25.7 & 6.15 & 16.7 - 49.5 & 29.3 & 5.43 & 12.0 - 41.8 & 23.5 & 5.49 \\
\textbf{Perimeter (Worst)} & mm & 50.4 - 251.2 & 107.3 & 33.6 & 85.1 - 251.2 & 141.4 & 29.5 & 50.4 - 127.1 & 87.0 & 13.5 \\
\textbf{Area (Worst)} & mm$^2$ & 185.2 - 4254.0 & 880.6 & 569.4 & 508.1 - 4254.0 & 1422.3 & 598.0 & 185.2 - 1210.0 & 558.9 & 163.6 \\
\textbf{Smoothness (Worst)} & ratio & 0.070 - 0.220 & 0.130 & 0.020 & 0.090 - 0.220 & 0.140 & 0.020 & 0.070 - 0.200 & 0.120 & 0.020 \\
\textbf{Compactness (Worst)} & ratio & 0.030 - 1.06 & 0.250 & 0.160 & 0.050 - 1.06 & 0.370 & 0.170 & 0.030 - 0.580 & 0.180 & 0.090 \\
\textbf{Concavity (Worst)} & ratio & 0.000 - 1.25 & 0.270 & 0.210 & 0.020 - 1.17 & 0.450 & 0.180 & 0.000 - 1.25 & 0.170 & 0.140 \\
\textbf{Concave Points (Worst)} & ratio & 0.000 - 0.290 & 0.110 & 0.070 & 0.030 - 0.290 & 0.180 & 0.050 & 0.000 - 0.170 & 0.070 & 0.040 \\
\textbf{Symmetry (Worst)} & ratio & 0.160 - 0.660 & 0.290 & 0.060 & 0.160 - 0.660 & 0.320 & 0.070 & 0.160 - 0.420 & 0.270 & 0.040 \\
\textbf{Fractal Dimension (Worst)} &  & 0.060 - 0.210 & 0.080 & 0.020 & 0.060 - 0.210 & 0.090 & 0.020 & 0.060 - 0.150 & 0.080 & 0.010 \\

\hline
\end{tabular}

\end{adjustbox}
\end{table}
\clearpage

%% file: table_3.tex
\renewcommand{\arraystretch}{1.5}
\begin{table}[tbp]

\caption{This table provides descriptive statistics for the features of the Heart Failure Prediction Dataset. The summary includes the range (minimum and maximum values), mean, and standard deviation (Std Dev) for continuous features, as well as counts and percentages for categorical features, across the entire dataset and stratified by heart disease status. Cholesterol and resting blood pressure are recorded as $0$ for 172 and 1 records respectively; because neither value is physiologically possible, these entries are treated as missing and excluded from the corresponding statistics.}

\centering
\begin{adjustbox}{max width=\textwidth}
\begin{tabular}{lllccccccccc}
\hline
\multicolumn{2}{c}{\multirow{4}{*}{\textbf{Risk Factor}}} & \multicolumn{1}{c}{\multirow{4}{*}{\textbf{Unit}}} & \multicolumn{9}{c}{\textbf{Continuous features}} \\ \cmidrule(lr){4-12} 
\multicolumn{2}{c}{} & \multicolumn{1}{c}{} & \multicolumn{3}{c}{\multirow{2}{*}{\textbf{All Population (\textit{N = 918})}}} & \multicolumn{6}{c}{\textbf{Heart Disease Status}} \\ \cmidrule(lr){7-12} 
\multicolumn{2}{c}{} & \multicolumn{1}{c}{} & \multicolumn{3}{c}{} & \multicolumn{3}{c}{\textbf{Disease \textit{(N = 508)}}} & \multicolumn{3}{c}{\textbf{No Disease\textit{ (N = 410)}}} \\ \cmidrule(lr){4-12} 
\multicolumn{2}{c}{} & \multicolumn{1}{c}{} & \textbf{Range} & \textbf{Mean} & \textbf{Std Dev} & \textbf{Range} & \textbf{Mean} & \textbf{Std Dev} & \textbf{Range} & \textbf{Mean} & \textbf{Std Dev} \\ \hline
\multicolumn{2}{l}{\textbf{Age}} & years & 28.0 - 77.0 & 53.5 & 9.43 & 31.0 - 77.0 & 55.9 & 8.73 & 28.0 - 76.0 & 50.6 & 9.44 \\
\multicolumn{2}{l}{\textbf{Resting Systolic Blood Pressure}} & mmHg & 80.0 - 200.0 & 132.5 & 18.0 & 92.0 - 200.0 & 134.4 & 18.9 & 80.0 - 190.0 & 130.2 & 16.5 \\
\multicolumn{2}{l}{\textbf{Cholesterol}} & mg/dL & 85.0 - 603.0 & 244.6 & 59.2 & 100.0 - 603.0 & 251.1 & 62.5 & 85.0 - 564.0 & 238.8 & 55.4 \\
\multicolumn{2}{l}{\textbf{Maximum Heart Rate}} & bpm & 60.0 - 202.0 & 136.8 & 25.5 & 60.0 - 195.0 & 127.7 & 23.4 & 69.0 - 202.0 & 148.2 & 23.3 \\
\multicolumn{2}{l}{\textbf{Oldpeak}} & mm & -2.60 - 6.20 & 0.890 & 1.07 & -2.60 - 6.20 & 1.27 & 1.15 & -1.10 - 4.20 & 0.410 & 0.700 \\
\hline
\multicolumn{2}{l}{\multirow{2}{*}{}} & \multirow{2}{*}{\textbf{Unit}} & \multicolumn{9}{c}{\textbf{Categorical features}} \\ \cmidrule(lr){4-12} 
\multicolumn{2}{l}{} & & \multicolumn{3}{c}{\textbf{All Population}} & \multicolumn{3}{c}{\textbf{Disease}} & \multicolumn{3}{c}{\textbf{No Disease}} \\ \hline

\multicolumn{2}{l}{\textbf{Sex}} & count  & \multicolumn{9}{l}{} \\
 & Male & & \multicolumn{3}{c}{725 (79.0\%)} & \multicolumn{3}{c}{458 (90.2\%)} & \multicolumn{3}{c}{267 (65.1\%)} \\
 & Female & & \multicolumn{3}{c}{193 (21.0\%)} & \multicolumn{3}{c}{50 (9.8\%)} & \multicolumn{3}{c}{143 (34.9\%)} \\
\multicolumn{2}{l}{\textbf{Chest Pain Type}} & count & \multicolumn{9}{l}{} \\
 & Atypical Angina & & \multicolumn{3}{c}{173 (18.8\%)} & \multicolumn{3}{c}{24 (4.7\%)} & \multicolumn{3}{c}{149 (36.3\%)} \\
 & Non-Anginal Pain & & \multicolumn{3}{c}{203 (22.1\%)} & \multicolumn{3}{c}{72 (14.2\%)} & \multicolumn{3}{c}{131 (32.0\%)} \\
 & Asymptomatic & & \multicolumn{3}{c}{496 (54.0\%)} & \multicolumn{3}{c}{392 (77.2\%)} & \multicolumn{3}{c}{104 (25.4\%)} \\
 & Typical Angina & & \multicolumn{3}{c}{46 (5.0\%)} & \multicolumn{3}{c}{20 (3.9\%)} & \multicolumn{3}{c}{26 (6.3\%)} \\
 \multicolumn{2}{l}{\textbf{Fasting Blood Sugar}} & count  & \multicolumn{9}{l}{} \\
 & No & & \multicolumn{3}{c}{704 (76.7\%)} & \multicolumn{3}{c}{338 (66.5\%)} & \multicolumn{3}{c}{366 (89.3\%)} \\
 & Yes & & \multicolumn{3}{c}{214 (23.3\%)} & \multicolumn{3}{c}{170 (33.5\%)} & \multicolumn{3}{c}{44 (10.7\%)} \\
\multicolumn{2}{l}{\textbf{Resting ECG}} & count  & \multicolumn{9}{l}{} \\
 & Normal & & \multicolumn{3}{c}{552 (60.1\%)} & \multicolumn{3}{c}{285 (56.1\%)} & \multicolumn{3}{c}{267 (65.1\%)} \\
 & ST & & \multicolumn{3}{c}{178 (19.4\%)} & \multicolumn{3}{c}{117 (23.0\%)} & \multicolumn{3}{c}{61 (14.9\%)} \\
 & LVH & & \multicolumn{3}{c}{188 (20.5\%)} & \multicolumn{3}{c}{106 (20.9\%)} & \multicolumn{3}{c}{82 (20.0\%)} \\
\multicolumn{2}{l}{\textbf{Exercise Angina}} & count  & \multicolumn{9}{l}{} \\
 & No & & \multicolumn{3}{c}{547 (59.6\%)} & \multicolumn{3}{c}{192 (37.8\%)} & \multicolumn{3}{c}{355 (86.6\%)} \\
 & Yes & & \multicolumn{3}{c}{371 (40.4\%)} & \multicolumn{3}{c}{316 (62.2\%)} & \multicolumn{3}{c}{55 (13.4\%)} \\
 \multicolumn{2}{l}{\textbf{ST Slope}} & count  & \multicolumn{9}{l}{} \\
 & Up & & \multicolumn{3}{c}{395 (43.0\%)} & \multicolumn{3}{c}{78 (15.4\%)} & \multicolumn{3}{c}{317 (77.3\%)} \\
 & Flat & & \multicolumn{3}{c}{460 (50.1\%)} & \multicolumn{3}{c}{381 (75.0\%)} & \multicolumn{3}{c}{79 (19.3\%)} \\
 & Down & & \multicolumn{3}{c}{63 (6.9\%)} & \multicolumn{3}{c}{49 (9.6\%)} & \multicolumn{3}{c}{14 (3.4\%)} \\
\hline
\end{tabular}

\end{adjustbox}
\label{tab:heart_dataset}
\end{table}
\clearpage

%% file: section_04_results.tex
\section{Results}
\label{sec:Results}

This section reports the comparative performance of the proposed Bernoulli Naïve Bayes framework on the three benchmark clinical datasets defined in Section~\ref{subsec:datasets_preprocessing}. Methodological details on preprocessing, thresholding, cross-validation, and statistical testing are provided in Section~\ref{sec:Methods}.

\subsection{Proposed Model Compared to Reference Methods}

The proposed $\chi^2$-based binarization was compared with alternative thresholding strategies within the same Bernoulli Naïve Bayes framework: mutual information, Gini gain, information gain, AUC-split, Otsu, median, mean, Youden's \(J\), and k-means midpoint.

Table~\ref{tab:binarization_variants_three_datasets} reports AUC (95\% CI) and pairwise DeLong/McNemar comparisons versus $\chi^2$. No alternative showed statistically significant improvement over $\chi^2$ in any dataset.

For the Pima Indians Diabetes Database, no statistically significant difference from $\chi^2$ was detected for any variant under either test. For the Wisconsin Breast Cancer dataset, $\chi^2$ and Gini gain achieved the highest AUC (0.984), while several alternatives underperformed in DeLong; McNemar identified only median thresholding as significantly worse than $\chi^2$. For the Heart Failure Prediction dataset, mutual information, Gini gain, and information gain matched $\chi^2$ (AUC 0.919), whereas Otsu, median, Youden's \(J\), and k-means midpoint underperformed the top methods according to both DeLong and McNemar tests.

Although $\chi^2$ was not always the single highest-performing method, it demonstrated the most consistent performance among the three datasets. Some alternatives were equal or numerically better in individual settings, but those same methods showed significant underperformance in others (for example, mutual information and information gain in Breast Cancer). Among the alternatives, Gini gain was the one for which significant differences from $\chi^2$ were detected least often.

\input{table_4}

\subsubsection{Pima Indians Diabetes Database}

Table \ref{tab:diabetes_statistical_comparison} reports the Pima Indians Diabetes Database comparisons for the preprocessing settings described in Section~\ref{subsec:datasets_preprocessing}. In these blocks, the self-implementations are generally aligned with the reported studies and frequently reproduce equal or higher values under the corresponding preprocessing protocols.

The statistical pattern in Table~\ref{tab:diabetes_statistical_comparison} is not one of uniform superiority over BNB. For most classical baselines no statistically significant difference from BNB was detected, with non-significant DeLong/McNemar outcomes in many rows. Significant advantages are concentrated in a limited subset, mainly neural/deep models and isolated variants (one random-forest block and selected non-linear SVM distance formulations), while significant underperformance appears repeatedly in tree-based baselines and some RBF-based variants.

For this dataset, the NaN-aware BNB implementation handles missing values internally by omitting likelihood contributions for missing entries rather than applying imputation. In most comparisons with standard BNB using $\chi^2$-based binarization, no statistically significant difference was detected, although NaN-aware BNB showed slightly higher values in most settings.

\subsubsection{Wisconsin Breast Cancer Dataset}

Table~\ref{tab:cancer_statistical_comparison} reports the Wisconsin Breast Cancer dataset comparisons without missing-value preprocessing, including BNB and the reference models from \cite{najiMachineLearningAlgorithms2021,jakharSELFStackedbasedEnsemble2024,shahMachineLearningTechniques2022,rovshenovPerformanceComparisonDifferent2022}. The implemented results in this table also show broad reproducibility, with many rows matching or surpassing the corresponding paper-level values.

Compared with the Pima Indians Diabetes Database and Heart Failure Prediction tables, Table~\ref{tab:cancer_statistical_comparison} contains a larger number of statistically significant outperformance cases versus BNB. The models that outperform BNB are mainly ensemble methods and neural/deep-learning models, plus selected SVM configurations with non-linear kernels, whereas several classical comparators remain non-significant and some tree-based models are significantly below BNB. These alternatives generally improve predictive performance at the cost of interpretability, because their complex and non-linear decision structures are less transparent for direct clinical reasoning.

\subsubsection{Heart Failure Prediction Dataset}

Table~\ref{tab:heart_statistical_comparison} reports Heart Failure Prediction dataset comparisons for BNB and the reference classifiers, including Gaussian Naïve Bayes \cite{muhammadEnhancingPrognosisAccuracy2023}. As in the other datasets, the independent implementations are generally consistent with the reported studies and in many rows achieve equal or higher computed metrics.

The statistical outcomes in Table~\ref{tab:heart_statistical_comparison} are predominantly non-significant for many classical baselines, so no difference from BNB was detected for these comparators. Significant outperformance is concentrated in selected ensemble-style configurations and isolated SVM cases, whereas repeated significant underperformance is observed for decision-tree baselines. This table reinforces the role of BNB as a competitive reference model and suggests that apparent SVM gains depending on non-linear kernel choices should be interpreted carefully, as such choices are less transparent for medical interpretation than linear kernel models.

\subsection{Calibration Analysis}

\begin{figure}[htbp]
    \centering
    \includegraphics[width=\textwidth,height=0.88\textheight,keepaspectratio]{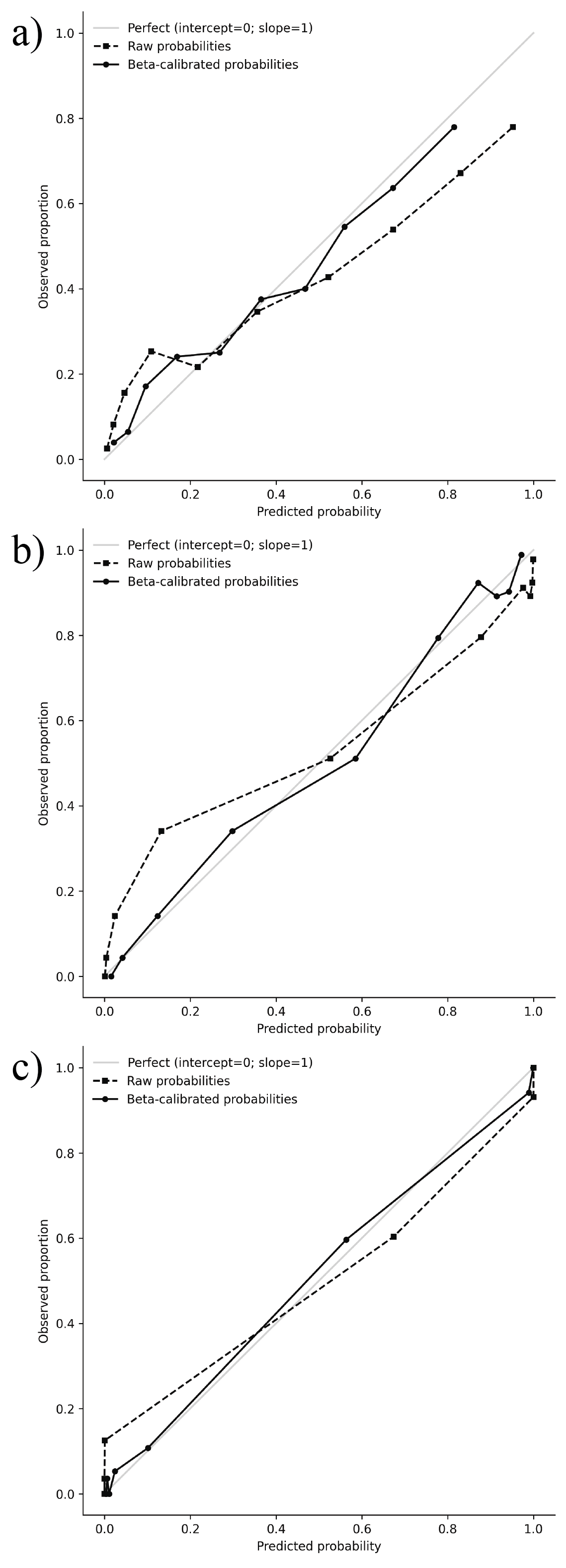}
    \caption{Calibration curves across three clinical datasets: (a) Pima Indians Diabetes Database, (b) Heart Failure Prediction dataset, and (c) Wisconsin Breast Cancer dataset. The diagonal line indicates perfect calibration. Dashed lines show raw probabilities and solid lines show three-parameter beta-calibrated probabilities.}
    \label{fig:calibration_plots}
\end{figure}

In addition to ROC-AUC, calibration analysis was performed following the protocol described in Section~\ref{sec:Methods}. Calibration results are summarized in Table~\ref{tab:calibration_performance} and the corresponding calibration curves are shown in Figure~\ref{fig:calibration_plots}. Four metrics are reported for both raw and beta-calibrated probabilities for each dataset: calibration-in-the-large, Brier score, calibration intercept, and calibration slope.

Raw BNB probabilities showed compressed slopes (ranging from 0.16 to 0.53), indicating overconfident probability estimates relative to the observed event rates. Beta calibration improved the slope substantially in all three datasets, with the Heart Failure dataset reaching a slope of 0.967 and an intercept of 0.006, close to ideal calibration. Brier scores also decreased after calibration in all cases. The Wisconsin Breast Cancer dataset had the lowest Brier score (0.043 post-calibration), consistent with its high discrimination (AUC 0.984).

\subsection{Explainability}
The interpretability of the proposed BNB framework is supported by complementary graphical and tabular analyses. Figures~\ref{fig:forest_pima_features}, \ref{fig:forest_wisconsin_features}, and \ref{fig:forest_heart_features} present feature-wise thresholds on a normalized $[0,1]$ scale together with the mean and mean $\pm$ standard deviation. This visualization makes the threshold location explicit relative to each feature distribution and clarifies how each predictor is partitioned into the Bernoulli states used for classification ($x \leq t$ and $x > t$).

For the Heart Failure dataset (Figure~\ref{fig:forest_heart_features}), binary predictors (Sex, Fasting Blood Sugar, and Exercise-Induced Angina) are separated at 0.5, while encoded categorical predictors (Resting ECG and ST Slope) are partitioned at category boundaries, yielding directly interpretable decision splits.

\input{table_5}
\input{table_6}
\input{table_7}

\input{table_8}

\input{figure_2}
\input{figure_3}
\input{figure_4}

In the Heart Failure dataset, categorical features were encoded for Categorical Naïve Bayes (CNB), as summarized in Table~\ref{tab:heart_dataset_encoding_summary}.

\input{table_9}

Tables~\ref{tab:pima_feature_summary}, \ref{tab:wisconsin_feature_summary}, and \ref{tab:heart_feature_summary_cnb} report outcome probabilities stratified by feature region, with 95\% confidence intervals obtained by bootstrap resampling ($n=1000$). For thresholded variables, these tables compare outcome probability at or below and above the threshold and therefore quantify both the direction and magnitude of feature effects on class membership. The absolute difference, $\Delta P$, is included as a practical proxy for class-separation strength, with larger values indicating stronger discrimination. For categorical rows in the Heart Failure CNB table, threshold-based $\Delta P$ is not applicable.

\input{table_10}
\input{table_11}
\input{table_12}

Perhaps the most direct demonstration of the framework's interpretability is that the entire classification procedure can be carried out by hand. Table~\ref{tab:pima_posterior_example} presents a complete leave-one-out worked example that integrates all explainability components. It shows how thresholds, conditional probabilities, and class priors are combined and normalized to classify a single sample. For each feature, the table reports the training-set count from which the conditional probability is obtained, so that every quantity can be checked independently, and the final rows give the product of the conditional probabilities, the effect of multiplying by the class prior, and the normalized posterior probabilities.

\input{table_13}

As shown in Table~\ref{tab:pima_posterior_example}, the sample is classified as Diabetes, with posterior probability 0.971, compared with 0.0295 for No Diabetes. This calculation does not include the usual logarithmic transformation of probabilities, which is commonly applied in Naïve Bayes implementations to avoid numerical underflow. This, however, does not affect the interpretability of the example, as the same terms are combined in the same way regardless of whether log transformation is applied.

%% file: table_4.tex
\begin{table}[tbp]
\centering
\caption{Binarization-variant sensitivity analysis for Bernoulli Naïve Bayes across the three datasets (10-fold cross-validation). For each dataset, pairwise DeLong and McNemar p-values are computed against the \texttt{chi2} baseline; $^{*}$ denotes significant improvement and $^{**}$ denotes significant underperformance.}
\label{tab:binarization_variants_three_datasets}
\begin{adjustbox}{max width=\textwidth}
\scriptsize
\begin{tabular}{@{}lccccccccc@{}}
\hline
\multicolumn{1}{c}{} & \multicolumn{3}{c}{\textbf{Pima Indians Diabetes}} & \multicolumn{3}{c}{\textbf{Wisconsin Breast Cancer}} & \multicolumn{3}{c}{\textbf{Heart Failure}} \\
\cline{2-10}
\textbf{Variant} & \textbf{AUC [95\% CI]} & \textbf{DeLong p} & \textbf{McNemar p} & \textbf{AUC [95\% CI]} & \textbf{DeLong p} & \textbf{McNemar p} & \textbf{AUC [95\% CI]} & \textbf{DeLong p} & \textbf{McNemar p} \\
\hline
chi2              & 0.794 [0.760, 0.826] & -      & -      & 0.984 [0.974, 0.992] & -      & -      & 0.919 [0.901, 0.936] & -      & -      \\
mutual\_info      & 0.799 [0.767, 0.830] & 0.3853 & 0.4510 & 0.981 [0.970, 0.990] & 0.0076$^{**}$ & 0.3877 & 0.919 [0.901, 0.936] & 0.3688 & 1.0000 \\
gini\_gain        & 0.792 [0.759, 0.824] & 0.4218 & 1.0000 & 0.984 [0.974, 0.992] & 0.2876 & 1.0000 & 0.919 [0.901, 0.936] & 1.0000 & 1.0000 \\
information\_gain & 0.799 [0.767, 0.830] & 0.3853 & 0.4510 & 0.981 [0.970, 0.990] & 0.0076$^{**}$ & 0.3877 & 0.919 [0.901, 0.936] & 0.3688 & 1.0000 \\
auc\_split        & 0.790 [0.759, 0.823] & 0.5187 & 0.7518 & 0.979 [0.968, 0.989] & 0.0468$^{**}$ & 0.1698 & 0.918 [0.900, 0.935] & 0.2595 & 1.0000 \\
otsu              & 0.780 [0.745, 0.810] & 0.3044 & 0.6276 & 0.983 [0.973, 0.991] & 0.6904 & 0.2863 & 0.908 [0.888, 0.925] & 0.0059$^{**}$ & 0.0865 \\
median            & 0.787 [0.755, 0.821] & 0.5024 & 0.2193 & 0.972 [0.958, 0.983] & 0.0102$^{**}$ & 0.0280$^{**}$ & 0.910 [0.890, 0.929] & 0.0010$^{**}$ & 0.0148$^{**}$ \\
mean              & 0.781 [0.748, 0.812] & 0.2056 & 0.4517 & 0.982 [0.971, 0.991] & 0.3718 & 0.2301 & 0.917 [0.899, 0.934] & 0.3624 & 0.5563 \\
youden\_j         & 0.790 [0.759, 0.823] & 0.5187 & 0.7518 & 0.978 [0.966, 0.989] & 0.0133$^{**}$ & 0.5413 & 0.859 [0.834, 0.882] & 0.0000$^{**}$ & 0.0000$^{**}$ \\
kmeans\_midpoint  & 0.775 [0.740, 0.808] & 0.0858 & 0.2807 & 0.972 [0.958, 0.985] & 0.0690 & 0.3833 & 0.908 [0.888, 0.926] & 0.0063$^{**}$ & 0.1273 \\
\hline
\end{tabular}
\end{adjustbox}
\end{table}

%% file: table_5.tex
\begin{table}[H]
\caption{Comparison of reported and reproduced algorithms for the Pima Indians Diabetes dataset under the preprocessing strategies used in the cited studies. Reported study values are shown alongside the corresponding implementations and their computed metrics. DeLong and McNemar p-values are calculated against the BNB benchmark row within each preprocessing block; bold p-values indicate statistical significance ($p<0.05$), with $^{*}$ denoting significant outperformance over BNB and $^{**}$ denoting significant underperformance. Only AUC is shown.\label{tab:diabetes_statistical_comparison}}
\begin{adjustwidth}{-\extralength}{0cm}
\centering
\setlength{\tabcolsep}{3pt}
\footnotesize
\begin{tabular}{@{}>{\centering\arraybackslash}p{1.7cm}>{\raggedright\arraybackslash}p{2.3cm}>{\raggedright\arraybackslash}p{3.9cm}>{\centering\arraybackslash}p{1.4cm}>{\centering\arraybackslash}p{3.3cm}|>{\centering\arraybackslash}p{1.4cm}>{\centering\arraybackslash}p{1.6cm}@{}}
\toprule
\multirow{2}{*}[-3.3pt]{\textbf{Reference}} & \multirow{2}{*}[-3.3pt]{\textbf{Preprocessing}} & \multirow{2}{*}[-3.3pt]{\textbf{Algorithm}} & \multicolumn{2}{c|}{\textbf{AUC}} & \multicolumn{2}{c}{\textbf{Statistical tests}} \\
\cmidrule(lr){4-5}\cmidrule(l){6-7}
 &  &  & \textbf{Reported} & \textbf{Reproduced [95\% CI]} & \textbf{DeLong} & \textbf{McNemar} \\
\midrule
\multirow{8}{*}{\cite{ashourOptimizedNeuralNetworks2024}} & \multirow{7}{2.3cm}{Mean imputation} & Gaussian Naïve Bayes & 0.846 & 0.820 [0.790,0.850] & 0.109 & 0.363 \\
 &  & k-Nearest Neighbors & 0.838 & 0.780 [0.740,0.810] & 0.352 & 0.399 \\
 &  & J48 Decision Tree & 0.785 & 0.710 [0.680,0.750] & \textbf{0.000$^{**}$} & 0.372 \\
 &  & Random Forest & 0.836 & 0.830 [0.790,0.850] & \textbf{0.002$^{*}$} & 0.648 \\
 &  & Feedforward Neural Network & 0.851 & 0.840 [0.810,0.860] & \textbf{0.000$^{*}$} & 0.176 \\
 &  & Convolutional Neural Network & 0.837 & 0.830 [0.810,0.860] & \textbf{0.000$^{*}$} & \textbf{0.030$^{*}$} \\
 &  & Bernoulli Naïve Bayes & -- & 0.790 [0.760,0.830] & 0.053 & \textbf{0.045$^{**}$} \\
 & - & Bernoulli Naïve Bayes (NaN-aware) & -- & 0.800 [0.770,0.840] & - & - \\
\midrule
\multirow{5}{*}{\cite{battineniComparativeMachineLearningApproach2019}} & \multirow{5}{2.3cm}{Removal of missing values} & Gaussian Naïve Bayes & 0.810 & 0.830 [0.790,0.870] & 0.793 & 0.419 \\
 &  & J48 Decision Tree & 0.750 & 0.800 [0.750,0.850] & 0.238 & 1.00 \\
 &  & Random Forest & 0.810 & 0.850 [0.810,0.900] & 0.062 & 0.896 \\
 &  & Logistic Regression & 0.830 & 0.850 [0.810,0.890] & 0.164 & 0.795 \\
 &  & Bernoulli Naïve Bayes & -- & 0.830 [0.780,0.870] & - & - \\
\midrule
\multirow{9}{*}{\cite{bhoi2021prediction}} & \multirow{9}{2.3cm}{Removal of missing values} & Classification Tree & 0.648 & 0.699 [0.649,0.749] & \textbf{0.000$^{**}$} & \textbf{0.015$^{**}$} \\
 &  & Support Vector Machine & 0.707 & 0.836 [0.790,0.878] & 0.592 & 0.200 \\
 &  & k-Nearest Neighbors & 0.737 & 0.797 [0.750,0.845] & 0.143 & 0.314 \\
 &  & Gaussian Naïve Bayes & 0.818 & 0.831 [0.790,0.872] & 0.793 & 0.419 \\
 &  & Random Forest & 0.808 & 0.851 [0.811,0.892] & 0.086 & 0.885 \\
 &  & Multilayer Perceptron & 0.824 & 0.804 [0.752,0.851] & 0.257 & 0.188 \\
 &  & AdaBoost & 0.681 & 0.843 [0.798,0.886] & 0.229 & 0.779 \\
 &  & Logistic Regression & 0.825 & 0.849 [0.808,0.890] & 0.153 & 0.896 \\
 &  & Bernoulli Naïve Bayes & -- & 0.830 [0.780,0.870] & -- & -- \\
\midrule
\multirow{6}{*}{\cite{prantoEvaluatingMachineLearning2020}} & \multirow{5}{2.3cm}{Mean imputation, feature selection} & Gaussian Naïve Bayes & 0.800 & 0.800 [0.770,0.840] & 0.687 & 0.593 \\
 &  & Decision Tree & 0.710 & 0.760 [0.720,0.790] & \textbf{0.020$^{**}$} & 0.350 \\
 &  & Random Forest & 0.830 & 0.810 [0.780,0.840] & 0.120 & 0.106 \\
 &  & k-Nearest Neighbors & 0.800 & 0.840 [0.810,0.860] & 0.150 & 0.317 \\
 &  & Bernoulli Naïve Bayes & -- & 0.790 [0.750,0.820] & 0.183 & 0.186 \\
 & - & Bernoulli Naïve Bayes (NaN-aware) & -- & 0.800 [0.770,0.830] & -- & -- \\
\midrule
\multirow{10}{*}{\cite{rezaImprovingSVMPerformance2023}} & \multirow{9}{2.3cm}{SMOTE, Mean imputation and Outlier Removal} & SVM (Linear) & 0.747 & 0.770 [0.740,0.810] & 0.312 & 0.539 \\
 &  & SVM (Polynomial) & 0.726 & 0.770 [0.740,0.810] & 0.303 & 0.344 \\
 &  & SVM (RBF Minkowski) & 0.836 & 0.740 [0.700,0.770] & \textbf{0.001$^{**}$} & 0.103 \\
 &  & SVM (RBF City Block) & 0.841 & 0.730 [0.700,0.770] & \textbf{0.001$^{**}$} & 0.062 \\
 &  & SVM (RBF Mahalanobis) & 0.817 & 0.790 [0.750,0.820] & 0.715 & 0.822 \\
 &  & SVM (RBF Bray-Curtis) & 0.744 & 0.810 [0.780,0.840] & 0.084 & 0.245 \\
 &  & SVM (RBF Canberra) & 0.789 & 0.830 [0.800,0.860] & \textbf{0.000$^{*}$} & 0.214 \\
 &  & SVM (RBF Euclidean Distance) & 0.809 & 0.830 [0.800,0.860] & \textbf{0.007$^{*}$} & 0.681 \\
 &  & Bernoulli Naïve Bayes & -- & 0.790 [0.750,0.820] & 0.178 & \textbf{0.016$^{**}$} \\
 & - & Bernoulli Naïve Bayes (NaN-aware) & -- & 0.800 [0.770,0.830] & -- & -- \\
\bottomrule
\end{tabular}
\end{adjustwidth}
\end{table}

%% file: table_6.tex
\begin{table}[H]
\caption{Comparison of reported and reproduced algorithms for the Wisconsin Breast Cancer dataset. Reported study values are shown alongside the corresponding implementations and their computed metrics. DeLong and McNemar p-values are calculated against the Bernoulli Naïve Bayes (BNB) benchmark within each study block; bold p-values indicate statistical significance ($p<0.05$), with $^{*}$ denoting significant outperformance over BNB and $^{**}$ denoting significant underperformance. Only AUC is shown.\label{tab:cancer_statistical_comparison}}
\begin{adjustwidth}{-\extralength}{0cm}
\centering
\setlength{\tabcolsep}{3pt}
\footnotesize
\begin{tabular}{@{}>{\centering\arraybackslash}p{1.7cm}>{\raggedright\arraybackslash}p{4.9cm}>{\centering\arraybackslash}p{1.4cm}>{\centering\arraybackslash}p{3.3cm}|>{\centering\arraybackslash}p{1.4cm}>{\centering\arraybackslash}p{1.6cm}@{}}
\toprule
\multirow{2}{*}[-3.3pt]{\textbf{Reference}} & \multirow{2}{*}[-3.3pt]{\textbf{Algorithm}} & \multicolumn{2}{c|}{\textbf{AUC}} & \multicolumn{2}{c}{\textbf{Statistical tests}} \\
\cmidrule(lr){3-4}\cmidrule(l){5-6}
 &  & \textbf{Reported} & \textbf{Reproduced [95\% CI]} & \textbf{DeLong} & \textbf{McNemar} \\
\midrule
\multirow{4}{*}{\cite{azarProbabilisticNeuralNetwork2013}} & Multi-layer Perceptron (MLP) & 0.993 & 0.992 [0.984,0.998] & 0.071 & \textbf{0.018$^{*}$} \\
 & Radial Basis Function Network (RBF) & 0.990 & 0.991 [0.983,0.997] & \textbf{0.033$^{*}$} & \textbf{0.007$^{*}$} \\
 & Probabilistic Neural Network (PNN) & 0.994 & 0.993 [0.987,0.998] & \textbf{0.015$^{*}$} & \textbf{0.002$^{*}$} \\
 & Bernoulli Naïve Bayes & -- & 0.984 [0.974,0.992] & - & - \\
\midrule
\multirow{11}{*}{\cite{jakharSELFStackedbasedEnsemble2024}} & SELF Stacking Ensemble & 0.991 & 0.992 [0.984,0.998] & \textbf{0.020$^{*}$} & \textbf{0.001$^{*}$} \\
 & Extra Trees & 0.994 & 0.993 [0.986,0.998] & \textbf{0.004$^{*}$} & \textbf{0.012$^{*}$} \\
 & Random Forest & 0.989 & 0.988 [0.978,0.996] & 0.071 & 0.607 \\
 & AdaBoost & 0.991 & 0.994 [0.989,0.998] & \textbf{0.004$^{*}$} & \textbf{0.001$^{*}$} \\
 & Gradient Boosting & 0.995 & 0.991 [0.983,0.996] & \textbf{0.021$^{*}$} & 0.115 \\
 & K-Nearest Neighbor (k=9) & 0.979 & 0.988 [0.978,0.997] & 0.276 & \textbf{0.018$^{*}$} \\
 & Support Vector Machine (RBF) & 0.963 & 0.994 [0.987,0.999] & \textbf{0.011$^{*}$} & \textbf{0.000$^{*}$} \\
 & Multilayer Perceptron (MLP) & 0.967 & 0.991 [0.982,0.998] & 0.147 & \textbf{0.009$^{*}$} \\
 & Classification and Regression Tree (CART) & 0.963 & 0.907 [0.882,0.932] & \textbf{0.000$^{**}$} & \textbf{0.009$^{**}$} \\
 & Stochastic Gradient Descent (Log-loss) & 0.951 & 0.994 [0.988,0.998] & \textbf{0.004$^{*}$} & \textbf{0.003$^{*}$} \\
 & Bernoulli Naïve Bayes & -- & 0.984 [0.974,0.992] & - & - \\
\midrule
\multirow{6}{*}{\cite{najiMachineLearningAlgorithms2021}} & Support Vector Machine (SVM) & 0.966 & 0.994 [0.987,0.999] & \textbf{0.011$^{*}$} & \textbf{0.000$^{*}$} \\
 & Random Forest & 0.960 & 0.989 [0.978,0.997] & 0.074 & 0.118 \\
 & Logistic Regression & 0.947 & 0.986 [0.977,0.995] & 0.214 & 0.157 \\
 & Decision Tree (C4.5 approximation) & 0.945 & 0.932 [0.909,0.954] & \textbf{0.000$^{**}$} & 0.499 \\
 & K-Nearest Neighbors (KNN) & 0.952 & 0.986 [0.975,0.996] & 0.586 & \textbf{0.016$^{*}$} \\
 & Bernoulli Naïve Bayes & -- & 0.984 [0.974,0.992] & - & - \\
\midrule
\multirow{4}{*}{\cite{rovshenovPerformanceComparisonDifferent2022}} & Artificial Neural Network (ANN) & 0.990 & 0.990 [0.982,0.996] & 0.099 & \textbf{0.018$^{*}$} \\
 & Support Vector Machine (SVM, RBF) & 0.970 & 0.995 [0.990,0.999] & \textbf{0.003$^{*}$} & \textbf{0.000$^{*}$} \\
 & Random Forest & 0.960 & 0.990 [0.980,0.997] & \textbf{0.027$^{*}$} & 0.210 \\
 & Bernoulli Naïve Bayes & -- & 0.984 [0.974,0.992] & - & - \\
\midrule
\multirow{4}{*}{\cite{shahMachineLearningTechniques2022}} & Decision Tree (Entropy) & 0.990 & 0.970 & \textbf{0.033$^{**}$} & 0.499 \\
 & Gaussian Naïve Bayes & 0.890 & 0.984 [0.975,0.992] & 0.878 & 0.167 \\
 & Random Forest & 0.990 & 0.989 [0.978,0.997] & 0.051 & 0.077 \\
 & Bernoulli Naïve Bayes & -- & 0.984 [0.974,0.992] & - & - \\
\bottomrule
\end{tabular}
\end{adjustwidth}
\end{table}

%% file: table_7.tex
\begin{table}[H]
\caption{Comparison of reported and reproduced algorithms for the Heart Failure Prediction dataset. Reported study values are shown alongside the corresponding implementations and their computed metrics. DeLong and McNemar p-values are calculated against the Bernoulli Naïve Bayes (BNB) benchmark within each study block; bold p-values indicate statistical significance ($p<0.05$), with $^{*}$ denoting significant outperformance over BNB and $^{**}$ denoting significant underperformance. Only AUC is shown.\label{tab:heart_statistical_comparison}}
\begin{adjustwidth}{-\extralength}{0cm}
\centering
\setlength{\tabcolsep}{3pt}
\footnotesize
\begin{tabular}{@{}>{\centering\arraybackslash}p{1.7cm}>{\raggedright\arraybackslash}p{4.9cm}>{\centering\arraybackslash}p{1.4cm}>{\centering\arraybackslash}p{3.3cm}|>{\centering\arraybackslash}p{1.4cm}>{\centering\arraybackslash}p{1.6cm}@{}}
\toprule
\multirow{2}{*}[-3.3pt]{\textbf{Reference}} & \multirow{2}{*}[-3.3pt]{\textbf{Algorithm}} & \multicolumn{2}{c|}{\textbf{AUC}} & \multicolumn{2}{c}{\textbf{Statistical tests}} \\
\cmidrule(lr){3-4}\cmidrule(l){5-6}
 &  & \textbf{Reported} & \textbf{Reproduced [95\% CI]} & \textbf{DeLong} & \textbf{McNemar} \\
\midrule
\multirow{7}{*}{\cite{kumarmygapulaPerformanceEvaluationMachine2023}} & SVM & 0.826 & 0.921 [0.902,0.940] & 0.901 & \textbf{0.029$^{*}$} \\
 & Random Forest & 0.838 & 0.929 [0.911,0.947] & \textbf{0.046$^{*}$} & \textbf{0.000$^{*}$} \\
 & Naïve Bayes & 0.853 & 0.911 [0.891,0.930] & 0.060 & 0.306 \\
 & Logistic Regression & 0.832 & 0.840 [0.815,0.864] & 0.253 & 0.418 \\
 & KNN & 0.827 & 0.910 [0.887,0.929] & 0.120 & 0.114 \\
 & Decision Tree & 0.805 & 0.798 [0.771,0.823] & \textbf{0.000$^{**}$} & \textbf{0.002$^{**}$} \\
 & Bernoulli Naïve Bayes & -- & 0.921 [0.901,0.938] & - & - \\
\midrule
\multirow{7}{*}{\cite{liuPredictiveClassifierCardiovascular2022}} & Logistic Regression (LR) & 0.940 & 0.911 [0.891,0.930] & \textbf{0.020$^{**}$} & 0.221 \\
 & Random Forest (RF) & 0.940 & 0.928 [0.910,0.945] & 0.126 & \textbf{0.012$^{*}$} \\
 & Extra Trees (ET) & 0.940 & 0.926 [0.908,0.944] & 0.357 & 0.075 \\
 & GBDT & 0.940 & 0.922 [0.903,0.940] & 0.980 & 0.470 \\
 & Multilayer Perceptron (MLP) & 0.920 & 0.923 [0.905,0.940] & 0.766 & 0.731 \\
 & Stacking Ensemble & 0.950 & 0.929 [0.911,0.946] & 0.092 & \textbf{0.011$^{*}$} \\
 & Bernoulli Naïve Bayes & -- & 0.922 [0.902,0.939] & - & - \\
\midrule
\multirow{7}{*}{\cite{muhammadEnhancingPrognosisAccuracy2023}} & Logistic Regression & 0.869 & 0.912 [0.891,0.930] & 0.120 & 1.00 \\
 & Decision Tree & 0.825 & 0.799 [0.773,0.824] & \textbf{0.000$^{**}$} & \textbf{0.003$^{**}$} \\
 & Random Forest & 0.866 & 0.927 [0.908,0.944] & 0.071 & 0.069 \\
 & Support Vector Machine (RBF) & 0.879 & 0.916 [0.894,0.935] & 0.521 & 0.101 \\
 & K Nearest Neighbors & 0.903 & 0.909 [0.887,0.927] & 0.069 & 1.00 \\
 & Gaussian Naïve Bayes & 0.863 & 0.909 [0.889,0.928] & 0.060 & 1.00 \\
 & Bernoulli Naïve Bayes & -- & 0.919 [0.901,0.936] & - & - \\
\midrule
\multirow{4}{*}{\cite{patidarComparativeAnalysisMachine2022}} & Random Forest & 0.990 & 0.930 [0.911,0.947] & \textbf{0.042$^{*}$} & \textbf{0.000$^{*}$} \\
 & Logistic Regression & 0.910 & 0.918 [0.897,0.936] & 0.421 & 0.418 \\
 & K-Nearest Neighbours (K=13) & 0.840 & 0.920 [0.901,0.939] & 0.983 & 0.282 \\
 & Bernoulli Naïve Bayes & -- & 0.921 [0.901,0.938] & - & - \\
\bottomrule
\end{tabular}
\end{adjustwidth}
\end{table}

%% file: table_8.tex
\begin{table}[tbp]
\centering
\caption{Calibration performance based on pooled out-of-fold predictions from 10-fold cross-validation. For each fold, beta calibration was fit on training-fold predictions and applied to the corresponding held-out fold. Ideal calibration intercept and slope are 0 and 1, respectively.}
\label{tab:calibration_performance}
\begin{adjustbox}{max width=\textwidth}
\begin{tabular}{lcccccc}
\hline
\multicolumn{1}{c}{\textbf{Dataset}} & \textbf{Model}  & \textbf{Brier} & \textbf{Intercept} & \textbf{Slope} & \textbf{Event rate} & \textbf{Mean risk} \\ \hline
\multirow{2}{*}{Pima Diabetes}       & Raw             & 0.182         & -0.281            & 0.529         & 0.349              & 0.373             \\
                                     & Beta-calibrated & 0.172         & -0.060            & 0.825         & 0.349              & 0.348             \\ \hline
\multirow{2}{*}{Breast Cancer}       & Raw             & 0.051         & -0.163            & 0.162         & 0.373              & 0.368             \\
                                     & Beta-calibrated & 0.043         & -0.091            & 0.843         & 0.373              & 0.370             \\ \hline
\multirow{2}{*}{Heart Disease}       & Raw             & 0.117         & 0.064             & 0.495         & 0.553              & 0.553             \\
                                     & Beta-calibrated & 0.108         & 0.006             & 0.967         & 0.553              & 0.554             \\ \hline
\end{tabular}

\end{adjustbox}
\end{table}

%% file: figure_2.tex
    \begin{figure}[htbp] 
        \hspace{-2cm} 
        \begin{tikzpicture}
            \begin{axis}[
                width=14cm, height=12cm,
                xlabel={Normalized scale (minimum value = 0, maximum value = 1)},
                ylabel={},
                ytick={1,2,3,4,5,6,7,8},
                y tick label style={align=right, text width=3.5cm, font=\footnotesize},
                yticklabels={{Age [21-81]}, {BMI [18.2-67.1]}, {Blood Pressure [24-122]}, {Diabetes Pedigree Function [0.078-2.42]}, {Glucose [44-199]}, {Insulin [14-846]}, {Pregnancies [0-17]}, {Skin Thickness [7-99]}},
                y dir=reverse,
                ymin=0.5, ymax=8.5,
                xmin= -0.1, xmax= 1.1,
                xtick={0, 0.2, 0.4, 0.6, 0.8, 1},
                xticklabels={Min, 20\%, 40\%, 60\%, 80\%, Max},
                axis x line=bottom,
                axis y line=left,
                legend style={at= [0.5, -0.15], anchor=north, cells={align=left,yshift=2pt}}
            ]
            \addplot[
                only marks,
                mark=square*,
                color=black,
                nodes near coords,
                point meta=explicit symbolic,
            every node near coord/.append style={font=\small,yshift=2pt}
            ] coordinates {
                           (0.20,1) [33.2]
                           (0.29,2) [32.5]
                           (0.49,3) [72.4]
                           (0.17,4) [0.472]
                           (0.50,5) [121.7]
                           (0.17,6) [155.5]
                           (0.23,7) [3.85]
                           (0.24,8) [29.2]
            };
            \addplot[thick, black] coordinates {(0.01,1) (0.40,1)};
            \addplot[thick, black] coordinates {(0.15,2) (0.43,2)};
            \addplot[thick, black] coordinates {(0.37,3) (0.62,3)};
            \addplot[thick, black] coordinates {(0.03,4) (0.31,4)};
            \addplot[thick, black] coordinates {(0.30,5) (0.70,5)};
            \addplot[thick, black] coordinates {(0.03,6) (0.31,6)};
            \addplot[thick, black] coordinates {(0.03,7) (0.42,7)};
            \addplot[thick, black] coordinates {(0.13,8) (0.35,8)};
    
            \addplot[
                only marks,
                mark=*,
                color=black,
                nodes near coords,
                point meta=explicit symbolic,
            every node near coord/.append style={font=\small,yshift=2pt}
            ] coordinates {
                (0.01,1) [21.5]
                (0.15,2) [25.5]
                (0.37,3) [60.0]
                (0.03,4) [0.141]
                (0.30,5) [91.2]
                (0.03,6) [36.8]
                (0.03,7) [0.475]
                (0.13,8) [18.7]
            };
            \addplot[only marks,
            mark=*,
            color=black,
            nodes near coords,
            point meta=explicit symbolic,
            every node near coord/.append style={font=\small,yshift=2pt}
            ] coordinates {
                (0.40,1) [45.0]
                (0.43,2) [39.4]
                (0.62,3) [84.8]
                (0.31,4) [0.803]
                (0.70,5) [152.2]
                (0.31,6) [274.3]
                (0.42,7) [7.21]
                (0.35,8) [39.6]
            };
            \addplot[
                only marks,
                mark=x,
                color=black,
                mark size=5,
                nodes near coords,
                point meta=explicit symbolic,
                every node near coord/.append style={anchor=north, font=\small,yshift=-2pt}
            ] coordinates {
                (0.12,1) [28.0]
                (0.24,2) [29.8]
                (0.45,3) [68.0]
                (0.19,4) [0.527]
                (0.54,5) [127.0]
                (0.13,6) [120.0]
                (0.35,7) [6.00]
                (0.17,8) [23.0]
            };
            \addplot[
                only marks,
                mark=square*,
                color=black
            ] coordinates {(0,0)};
            \addplot[thick, black] coordinates {(0,0) (1,0)};
            \addplot[
                only marks,
                mark=*,
                color=black
            ] coordinates {(0,0)};
            \addplot[
                only marks,
                mark=x,
                color=black,
                mark size=5
            ] coordinates {(0,0)};
        \end{axis}
        \end{tikzpicture}
        
        \textbf{Legend:} 
        \textcolor{black}{$\blacksquare$} Mean, 
        \textcolor{black}{$\bullet$} $\pm$ Standard Deviation, 
        \textcolor{black}{$\times$} Binarization Threshold.
        
    \caption{Forest Plot of Pima Indian Diabetes Database. The plot presents the distribution of key features on a normalized scale (0 to 1). The y-axis lists the features along with their respective value ranges, while the x-axis represents their normalized values. Horizontal lines indicate the spread of data, with markers representing key statistical values.
    }
    \label{fig:forest_pima_features}   
    \end{figure}

%% file: figure_3.tex
    \clearpage
    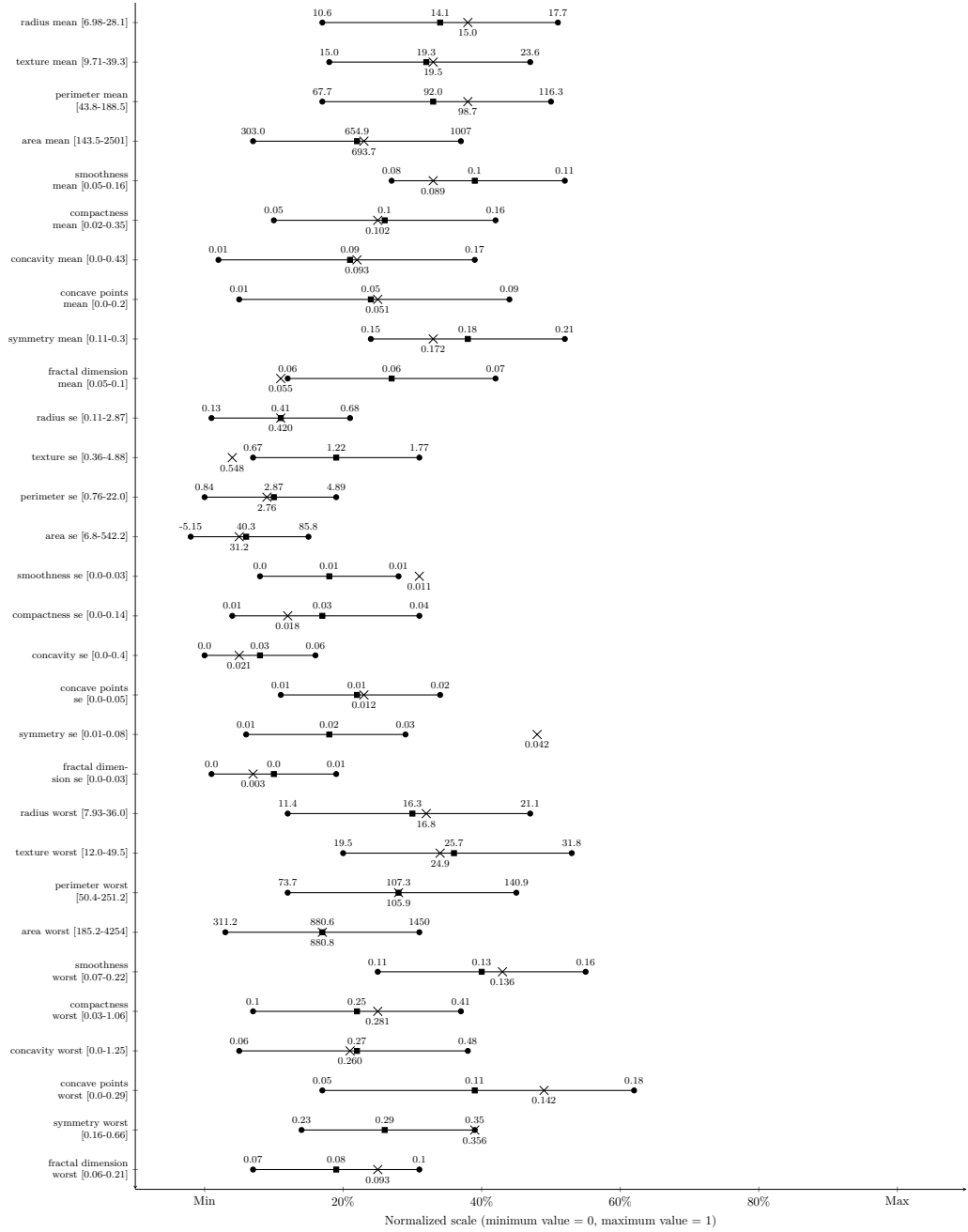
\begin{figure}[htbp] 
     \hspace{-2cm}
        \centering
        \scalebox{0.5}{
        \begin{tikzpicture}
            \begin{axis}[
                width=25cm, height=35cm,
                xlabel={Normalized scale (minimum value = 0, maximum value = 1)},
                ylabel={}, 
                ytick={1,2,3,4,5,6,7,8,9,10,11,12,13,14,15,16,17,18,19,20,21,22,23,24,25,26,27,28,29,30},
              y tick label style={align=right, text width=3.5cm, font=\footnotesize},
              yticklabels={{radius mean [6.98-28.1]}, {texture mean [9.71-39.3]}, {perimeter mean [43.8-188.5]}, {area mean [143.5-2501]}, {smoothness mean [0.05-0.16]}, {compactness mean [0.02-0.35]}, {concavity mean [0.0-0.43]}, {concave points mean [0.0-0.2]}, {symmetry mean [0.11-0.3]}, {fractal dimension mean [0.05-0.1]}, {radius se [0.11-2.87]}, {texture se [0.36-4.88]}, {perimeter se [0.76-22.0]}, {area se [6.8-542.2]}, {smoothness se [0.0-0.03]}, {compactness se [0.0-0.14]}, {concavity se [0.0-0.4]}, {concave points se [0.0-0.05]}, {symmetry se [0.01-0.08]}, {fractal dimension se [0.0-0.03]}, {radius worst [7.93-36.0]}, {texture worst [12.0-49.5]}, {perimeter worst [50.4-251.2]}, {area worst [185.2-4254]}, {smoothness worst [0.07-0.22]}, {compactness worst [0.03-1.06]}, {concavity worst [0.0-1.25]}, {concave points worst [0.0-0.29]}, {symmetry worst [0.16-0.66]}, {fractal dimension worst [0.06-0.21]}}, 
                y dir=reverse,
                ymin=0.5, ymax=30.5,
                xmin= -0.1, xmax= 1.1,
                xtick={0, 0.2, 0.4, 0.6, 0.8, 1},
                xticklabels={Min, 20\%, 40\%, 60\%, 80\%, Max},
                axis x line=bottom,
                axis y line=left,
                legend style={at= [0.5, -0.15], anchor=north, cells={align=left,yshift=2pt}}
            ]
            \addplot[
                only marks,
                mark=square*,
                color=black,
                nodes near coords,
                point meta=explicit symbolic,
            every node near coord/.append style={font=\footnotesize,yshift=2pt}
            ] coordinates {(0.34, 1) [14.1]
                           (0.32, 2) [19.3]
                           (0.33, 3) [92.0]
                           (0.22, 4) [654.9]
                           (0.39, 5) [0.1]
                           (0.26, 6) [0.1]
                           (0.21, 7) [0.09]
                           (0.24, 8) [0.05]
                           (0.38, 9) [0.18]
                           (0.27, 10) [0.06]
                           (0.11, 11) [0.41]
                           (0.19, 12) [1.22]
                           (0.1, 13) [2.87]
                           (0.06, 14) [40.3]
                           (0.18, 15) [0.01]
                           (0.17, 16) [0.03]
                           (0.08, 17) [0.03]
                           (0.22, 18) [0.01]
                           (0.18, 19) [0.02]
                           (0.1, 20) [0.0]
                           (0.3, 21) [16.3]
                           (0.36, 22) [25.7]
                           (0.28, 23) [107.3]
                           (0.17, 24) [880.6]
                           (0.4, 25) [0.13]
                           (0.22, 26) [0.25]
                           (0.22, 27) [0.27]
                           (0.39, 28) [0.11]
                           (0.26, 29) [0.29]
                           (0.19, 30) [0.08]
            };
    
            \addplot[thick, black] coordinates {(0.17,1) (0.51,1)};
            \addplot[thick, black] coordinates {(0.18,2) (0.47,2)};
            \addplot[thick, black] coordinates {(0.17,3) (0.5,3)};
            \addplot[thick, black] coordinates {(0.07,4) (0.37,4)};
            \addplot[thick, black] coordinates {(0.27,5) (0.52,5)};
            \addplot[thick, black] coordinates {(0.1,6) (0.42,6)};
            \addplot[thick, black] coordinates {(0.02,7) (0.39,7)};
            \addplot[thick, black] coordinates {(0.05,8) (0.44,8)};
            \addplot[thick, black] coordinates {(0.24,9) (0.52,9)};
            \addplot[thick, black] coordinates {(0.12,10) (0.42,10)};
            \addplot[thick, black] coordinates {(0.01,11) (0.21,11)};
            \addplot[thick, black] coordinates {(0.07,12) (0.31,12)};
            \addplot[thick, black] coordinates {(0.0,13) (0.19,13)};
            \addplot[thick, black] coordinates {(-0.02,14) (0.15,14)};
            \addplot[thick, black] coordinates {(0.08,15) (0.28,15)};
            \addplot[thick, black] coordinates {(0.04,16) (0.31,16)};
            \addplot[thick, black] coordinates {(0.0,17) (0.16,17)};
            \addplot[thick, black] coordinates {(0.11,18) (0.34,18)};
            \addplot[thick, black] coordinates {(0.06,19) (0.29,19)};
            \addplot[thick, black] coordinates {(0.01,20) (0.19,20)};
            \addplot[thick, black] coordinates {(0.12,21) (0.47,21)};
            \addplot[thick, black] coordinates {(0.2,22) (0.53,22)};
            \addplot[thick, black] coordinates {(0.12,23) (0.45,23)};
            \addplot[thick, black] coordinates {(0.03,24) (0.31,24)};
            \addplot[thick, black] coordinates {(0.25,25) (0.55,25)};
            \addplot[thick, black] coordinates {(0.07,26) (0.37,26)};
            \addplot[thick, black] coordinates {(0.05,27) (0.38,27)};
            \addplot[thick, black] coordinates {(0.17,28) (0.62,28)};
            \addplot[thick, black] coordinates {(0.14,29) (0.39,29)};
            \addplot[thick, black] coordinates {(0.07,30) (0.31,30)};
    
            \addplot[
                only marks,
                mark=*,
                color=black,
                nodes near coords,
                point meta=explicit symbolic,
            every node near coord/.append style={font=\footnotesize,yshift=2pt}
            ] coordinates {            (0.17,1) [10.6]
                (0.18,2) [15.0]
                (0.17,3) [67.7]
                (0.07,4) [303.0]
                (0.27,5) [0.08]
                (0.1,6) [0.05]
                (0.02,7) [0.01]
                (0.05,8) [0.01]
                (0.24,9) [0.15]
                (0.12,10) [0.06]
                (0.01,11) [0.13]
                (0.07,12) [0.67]
                (0.0,13) [0.84]
                (-0.02,14) [-5.15]
                (0.08,15) [0.0]
                (0.04,16) [0.01]
                (0.0,17) [0.0]
                (0.11,18) [0.01]
                (0.06,19) [0.01]
                (0.01,20) [0.0]
                (0.12,21) [11.4]
                (0.2,22) [19.5]
                (0.12,23) [73.7]
                (0.03,24) [311.2]
                (0.25,25) [0.11]
                (0.07,26) [0.1]
                (0.05,27) [0.06]
                (0.17,28) [0.05]
                (0.14,29) [0.23]
                (0.07,30) [0.07]
            };
            \addplot[only marks,
            mark=*,
            color=black,
            nodes near coords,
            point meta=explicit symbolic,
            every node near coord/.append style={font=\footnotesize,yshift=2pt}
            ] coordinates {
                (0.51,1) [17.7]
                (0.47,2) [23.6]
                (0.5,3) [116.3]
                (0.37,4) [1007]
                (0.52,5) [0.11]
                (0.42,6) [0.16]
                (0.39,7) [0.17]
                (0.44,8) [0.09]
                (0.52,9) [0.21]
                (0.42,10) [0.07]
                (0.21,11) [0.68]
                (0.31,12) [1.77]
                (0.19,13) [4.89]
                (0.15,14) [85.8]
                (0.28,15) [0.01]
                (0.31,16) [0.04]
                (0.16,17) [0.06]
                (0.34,18) [0.02]
                (0.29,19) [0.03]
                (0.19,20) [0.01]
                (0.47,21) [21.1]
                (0.53,22) [31.8]
                (0.45,23) [140.9]
                (0.31,24) [1450]
                (0.55,25) [0.16]
                (0.37,26) [0.41]
                (0.38,27) [0.48]
                (0.62,28) [0.18]
                (0.39,29) [0.35]
                (0.31,30) [0.1]
            };
            \addplot[
                only marks,
                mark=x,
                color=black,
                mark size=5,
                nodes near coords,
                point meta=explicit symbolic,
                every node near coord/.append style={anchor=north, font=\footnotesize,yshift=-2pt}
            ] coordinates {
                (0.38,1) [15.0]
                (0.33,2) [19.5]
                (0.38,3) [98.7]
                (0.23,4) [693.7]
                (0.33,5) [0.089]
                (0.25,6) [0.102]
                (0.22,7) [0.093]
                (0.25,8) [0.051]
                (0.33,9) [0.172]
                (0.11,10) [0.055]
                (0.11,11) [0.420]
                (0.04,12) [0.548]
                (0.09,13) [2.76]
                (0.05,14) [31.2]
                (0.31,15) [0.011]
                (0.12,16) [0.018]
                (0.05,17) [0.021]
                (0.23,18) [0.012]
                (0.48,19) [0.042]
                (0.07,20) [0.003]
                (0.32,21) [16.8]
                (0.34,22) [24.9]
                (0.28,23) [105.9]
                (0.17,24) [880.8]
                (0.43,25) [0.136]
                (0.25,26) [0.281]
                (0.21,27) [0.260]
                (0.49,28) [0.142]
                (0.39,29) [0.356]
                (0.25,30) [0.093]
            };
            \addplot[
                only marks,
                mark=square*,
                color=black
            ] coordinates {(0,0)};
            \addplot[thick, black] coordinates {(0,0) (1,0)};
            \addplot[
                only marks,
                mark=*,
                color=black
            ] coordinates {(0,0)};
            \addplot[
                only marks,
                mark=x,
                color=black,
                mark size=5
            ] coordinates {(0,0)};
        \end{axis}
        \end{tikzpicture}
        }
    
        \textbf{Legend:} 
        \textcolor{black}{$\blacksquare$} Mean, 
        \textcolor{black}{$\bullet$} $\pm$ Standard Deviation, 
        \textcolor{black}{$\times$} Binarization Threshold.
    
        \caption{Forest Plot of Wisconsin Breast Cancer Diagnosis dataset. This plot displays the range, mean, standard deviation, and optimal binarization threshold for each feature on a normalized scale (0 to 1). The binarization threshold is used to transform continuous medical data into a Bernoulli distribution, enabling the application of the Bernoulli Naïve Bayes classifier for supervised classification.}
        \label{fig:forest_wisconsin_features}
    \end{figure}
        

%% file: figure_4.tex
    \clearpage
    \begin{figure}[htbp] 
           \hspace{-2cm} 
        \centering
        \begin{tikzpicture}
            \begin{axis}[
                width=12cm, height=17cm,
                xlabel={Normalized scale (minimum value = 0, maximum value = 1)},
                ylabel={}, 
                ytick={1,2,3,4,5,6,7,8,9,10,11},
              y tick label style={align=right, text width=3.5cm, font=\footnotesize},
              yticklabels={{Age [28-77]}, {Sex [0-1]}, {Chest Pain Type [0-3]}, {Resting BP [80-200]}, {Cholesterol [85-603]}, {Fasting BS [0-1]}, {Resting ECG [0-2]}, {Max HR [60-202]}, {Exercise Angina [0-1]}, {Oldpeak [-2.6-6.2]}, {ST Slope [0-2]}},
                y dir=reverse,
                ymin=0.5, ymax=11.5,
                xmin= -0.1, xmax= 1.1,
                xtick={0, 0.2, 0.4, 0.6, 0.8, 1},
                xticklabels={Min, 20\%, 40\%, 60\%, 80\%, Max},
                axis x line=bottom,
                axis y line=left,
                legend style={at= [0.5, -0.15], anchor=north, cells={align=left,yshift=2pt}}
            ]
            \addplot[
                only marks,
                mark=square*,
                color=black,
                nodes near coords,
                point meta=explicit symbolic,
            every node near coord/.append style={font=\footnotesize,yshift=2pt}
            ] coordinates {(0.52, 1) [53.5]
                           (0.79, 2) [0.79]
                           (0.26, 3) [0.78]
                           (0.44, 4) [132.5]
                           (0.31, 5) [244.6]
                           (0.23, 6) [0.23]
                           (0.49, 7) [0.99]
                           (0.54, 8) [136.8]
                           (0.4, 9) [0.4]
                           (0.4, 10) [0.89]
                           (0.68, 11) [1.36]
            };
    
            \addplot[thick, black] coordinates {(0.33,1) (0.71,1)};
            \addplot[thick, black] coordinates {(0.38,2) (1.2,2)};
            \addplot[thick, black] coordinates {(-0.06,3) (0.58,3)};
            \addplot[thick, black] coordinates {(0.29,4) (0.59,4)};
            \addplot[thick, black] coordinates {(0.19,5) (0.42,5)};
            \addplot[thick, black] coordinates {(-0.19,6) (0.66,6)};
            \addplot[thick, black] coordinates {(0.18,7) (0.81,7)};
            \addplot[thick, black] coordinates {(0.36,8) (0.72,8)};
            \addplot[thick, black] coordinates {(-0.09,9) (0.9,9)};
            \addplot[thick, black] coordinates {(0.28,10) (0.52,10)};
            \addplot[thick, black] coordinates {(0.38,11) (0.98,11)};
    
            \addplot[
                only marks,
                mark=*,
                color=black,
                nodes near coords,
                point meta=explicit symbolic,
            every node near coord/.append style={font=\footnotesize,yshift=2pt}
            ] coordinates {            (0.33,1) [44.1]
                (0.38,2) [0.38]
                (-0.06,3) [-0.18]
                (0.29,4) [114.5]
                (0.19,5) [185.4]
                (-0.19,6) [-0.19]
                (0.18,7) [0.36]
                (0.36,8) [111.3]
                (-0.09,9) [-0.09]
                (0.28,10) [-0.18]
                (0.38,11) [0.75]
            };
            \addplot[only marks,
            mark=*,
            color=black,
            nodes near coords,
            point meta=explicit symbolic,
            every node near coord/.append style={font=\footnotesize,yshift=2pt}
            ] coordinates {
                (0.71,1) [62.9]
                (1.2,2) [1.2]
                (0.58,3) [1.74]
                (0.59,4) [150.5]
                (0.42,5) [303.8]
                (0.66,6) [0.66]
                (0.81,7) [1.62]
                (0.72,8) [162.3]
                (0.9,9) [0.9]
                (0.52,10) [1.95]
                (0.98,11) [1.97]
            };
            \addplot[
                only marks,
                mark=x,
                color=black,
                mark size=5,
                nodes near coords,
                point meta=explicit symbolic,
                every node near coord/.append style={anchor=north, font=\footnotesize,yshift=-2pt}
            ] coordinates {
                (0.53,1) [54.0]
                (0.5,2) [0.5]
                (0.33,3) [1]
                (0.50,4) [140.0]
                (0.32,5) [253.0]
                (0.5,6) [0.5]
                (1.0,7) [2]
                (0.51,8) [132.0]
                (0.5,9) [0.5]
                (0.36,10) [0.600]
                (1.0,11) [2]
            };
            \addplot[
                only marks,
                mark=square*,
                color=black
            ] coordinates {(0,0)};
            \addplot[thick, black] coordinates {(0,0) (1,0)};
            \addplot[
                only marks,
                mark=*,
                color=black
            ] coordinates {(0,0)};
            \addplot[
                only marks,
                mark=x,
                color=black,
                mark size=5
            ] coordinates {(0,0)};
        \end{axis}
        \end{tikzpicture}
        \textbf{Legend:} 
        \textcolor{black}{$\blacksquare$} Mean, 
        \textcolor{black}{$\bullet$} $\pm$ Standard Deviation, 
        \textcolor{black}{$\times$} Binarization Threshold.
        
    \caption{Forest Plot of Heart Failure Prediction Dataset. This plot presents the distribution of key features on a normalized scale (0 to 1). The y-axis lists the features along with their respective value ranges, while the x-axis represents their normalized values. Horizontal lines indicate the spread of data, with markers representing key statistical values.
    }
    \label{fig:forest_heart_features}   
    \end{figure}
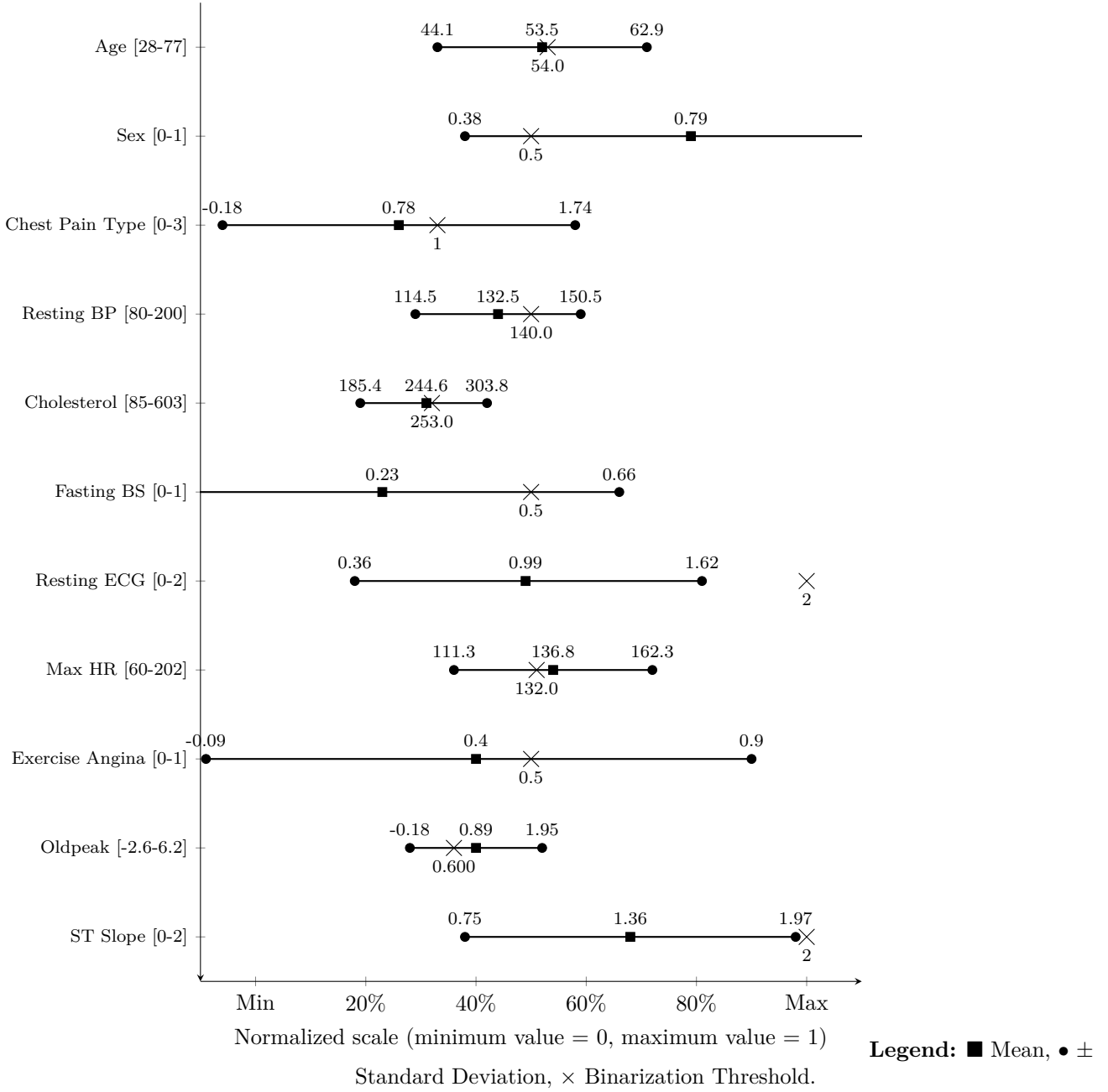

%% file: table_9.tex
\begin{table}[tbp]
    \centering
    \caption{Encoding table for the Heart Failure Prediction Dataset. This table provides the categorical features, their respective categories, and the encoded values used in the dataset.}
        \label{tab:heart_dataset_encoding_summary}
        \begin{adjustbox}{max width=\textwidth}
        \begin{tabular}{lcc}
        \hline
        \multicolumn{1}{c}{\textbf{Column}} & \textbf{Category} & \textbf{Encoded Value} \\ \hline
        \multirow{2}{*}{Sex}                & F                 & 0.000                      \\
                            & M                 & 1.00                      \\ \hline
        \multirow{4}{*}{Chest Pain Type}    & ASY               & 0.000                      \\
                            & ATA               & 1.00                      \\
                            & NAP               & 2.00                      \\
                            & TA                & 3.00                      \\ \hline
        \multirow{3}{*}{Resting ECG}        & LVH               & 0.000                      \\
                            & Normal            & 1.00                      \\
                            & ST                & 2.00                      \\ \hline
        \multirow{2}{*}{Exercise Angina}    & N                 & 0.000                      \\
                            & Y                 & 1.00                      \\ \hline
        \multirow{3}{*}{ST Slope}           & Down              & 0.000                      \\
                            & Flat              & 1.00                      \\
                            & Up                & 2.00                      \\ \hline
        \end{tabular}

        \end{adjustbox}
        \end{table}

%% file: table_10.tex
\renewcommand{\arraystretch}{1.5}    

\begin{table}[tbp]
    \caption{Summary statistics for PIMA dataset features. Here $t$ denotes the chi-square-selected threshold for each feature, $P(y{=}1 \mid x \leq t) \equiv P(\text{diabetes} \mid \text{feature} \leq \text{threshold})$ and $P(y{=}1 \mid x > t) \equiv P(\text{diabetes} \mid \text{feature} > \text{threshold})$ are the conditional probabilities of diabetes at or below and above the threshold, respectively, and $\Delta P$ is their absolute difference. Values in brackets denote 95\% confidence intervals estimated via bootstrap resampling ($n = 1000$).}
    \label{tab:pima_feature_summary}

    \begin{adjustbox}{max width=\textwidth}

    \begin{tabular}{lcccccc}
    \hline
    \textbf{Feature} & \textbf{Min} & \textbf{Max} & \textbf{$threshold$ {[}95\% CI{]}} & \textbf{$P(y{=}1 \mid x \leq t)$ {[}95\% CI{]}} & \textbf{$P(y{=}1 \mid x > t)$ {[}95\% CI{]}} & \textbf{$\Delta P$ {[}95\% CI{]}} \\ \hline
    
Pregnancies                & 0.000     & 17.0    & 6.00 [2.00, 7.00]       & 0.289 [0.212, 0.330]& 0.562 [0.370, 0.595]& 0.273 [0.104, 0.313] \\
Glucose                    & 44.0    & 199.0   & 127.0 [112.0, 159.0] & 0.192 [0.117, 0.276]& 0.615 [0.522, 0.854]& 0.423 [0.365, 0.599] \\
Blood Pressure             & 24.0    & 122.0   & 68.0 [60.0, 96.0]    & 0.247 [0.106, 0.338]& 0.404 [0.343, 0.602]& 0.157 [0.105, 0.312] \\
Skin Thickness             & 7.00     & 99.0    & 23.0 [19.0, 34.0]    & 0.157 [0.068, 0.272]& 0.415 [0.353, 0.533]& 0.258 [0.173, 0.342] \\
Insulin                    & 14.0    & 846.0   & 120.0 [87.0, 152.0]  & 0.146 [0.044, 0.206]& 0.505 [0.407, 0.594]& 0.359 [0.286, 0.474] \\
BMI                        & 18.2    & 67.1    & 29.8 [26.2, 31.2]    & 0.161 [0.041, 0.203]& 0.463 [0.392, 0.536]& 0.303 [0.262, 0.416] \\
Diabetes Pedigree Function & 0.078     & 2.42     & 0.527 [0.200, 1.10]       & 0.291 [0.148, 0.345]& 0.463 [0.370, 0.721]& 0.173 [0.144, 0.410] \\
Age                        & 21.0    & 81.0    & 28.0 [24.0, 31.0]    & 0.193 [0.107, 0.244]& 0.491 [0.404, 0.545]& 0.298 [0.225, 0.375] \\

    \hline
    \end{tabular}%

    \end{adjustbox}
\end{table}

%% file: table_11.tex
\renewcommand{\arraystretch}{1.5}    

\begin{table}[tbp]
    \caption{Summary statistics for Wisconsin Breast Cancer Diagnosis Dataset features. Here $t$ denotes the chi-square-selected threshold for each feature, $P(y{=}1 \mid x \leq t) \equiv P(\text{cancer} \mid \text{feature} \leq \text{threshold})$ and $P(y{=}1 \mid x > t) \equiv P(\text{cancer} \mid \text{feature} > \text{threshold})$ are the conditional probabilities of malignancy at or below and above the threshold, respectively, and $\Delta P$ is their absolute difference. Values in brackets denote 95\% confidence intervals estimated via bootstrap resampling ($n = 1000$).}
    \label{tab:wisconsin_feature_summary}

    \begin{adjustbox}{max width=\textwidth}

    \begin{tabular}{lcccccc}
    \hline
    \textbf{Feature} & \textbf{Min} & \textbf{Max} & \textbf{$t$ {[}95\% CI{]}} & \textbf{$P(y{=}1 \mid x \leq t)$ {[}95\% CI{]}} & \textbf{$P(y{=}1 \mid x > t)$ {[}95\% CI{]}} & \textbf{$\Delta P$ {[}95\% CI{]}} \\ \hline
    Radius (Mean)                  & 6.98  & 28.1  & 15.0 [14.4, 15.3]     & 0.129 [0.084, 0.170]& 0.931 [0.848, 0.965]& 0.802 [0.728, 0.852] \\
    Texture (Mean)                 & 9.71  & 39.3  & 19.5 [18.0, 20.2]     & 0.177 [0.087, 0.220]& 0.619 [0.531, 0.697]& 0.442 [0.380, 0.542] \\
    Perimeter (Mean)               & 43.8 & 188.5 & 98.7 [89.8, 99.6]     & 0.133 [0.053, 0.159]& 0.941 [0.807, 0.973]& 0.808 [0.728, 0.863] \\
    Area (Mean)                    & 143.5 & 2501.0 & 693.7 [641.2, 711.8] & 0.126 [0.083, 0.164]& 0.942 [0.855, 0.970]& 0.816 [0.732, 0.861] \\
    Smoothness (Mean)              & 0.053  & 0.163   & 0.089 [0.088, 0.108]        & 0.134 [0.068, 0.282]& 0.488 [0.441, 0.694]& 0.354 [0.292, 0.462] \\
    Compactness (Mean)             & 0.019  & 0.345   & 0.102 [0.095, 0.124]        & 0.118 [0.071, 0.211]& 0.686 [0.621, 0.840]& 0.568 [0.507, 0.677] \\
    Concavity (Mean)               & 0.000  & 0.427   & 0.093 [0.071, 0.111]        & 0.090 [0.035, 0.137]& 0.845 [0.759, 0.923]& 0.755 [0.693, 0.829] \\
    Concave Points (Mean)          & 0.000  & 0.201   & 0.051 [0.045, 0.058]        & 0.057 [0.026, 0.098]& 0.869 [0.815, 0.940]& 0.811 [0.759, 0.876] \\
    Symmetry (Mean)                & 0.106  & 0.304   & 0.172 [0.171, 0.211]        & 0.197 [0.126, 0.338]& 0.491 [0.450, 0.812]& 0.295 [0.261, 0.503] \\
    Fractal Dimension (Mean)       & 0.050  & 0.097   & 0.055 [0.052, 0.073]        & 0.630 [0.314, 1.000]& 0.350 [0.288, 0.661]& 0.281 [0.176, 0.658] \\
    Radius (SE)                    & 0.112  & 2.87   & 0.420 [0.372, 0.546]        & 0.164 [0.108, 0.233]& 0.801 [0.709, 0.944]& 0.637 [0.566, 0.736] \\
    Texture (SE)                   & 0.360  & 4.89   & 0.548 [0.531, 2.22]        & 0.097 [0.000, 0.442]& 0.388 [0.152, 0.467]& 0.292 [0.120, 0.413] \\
    Perimeter (SE)                 & 0.757  & 22.0  & 2.76 [2.40, 3.43]        & 0.148 [0.095, 0.216]& 0.775 [0.689, 0.887]& 0.627 [0.556, 0.720] \\
    Area (SE)                      & 6.80  & 542.2 & 31.2 [30.5, 42.8]     & 0.101 [0.068, 0.170]& 0.830 [0.793, 0.946]& 0.729 [0.687, 0.825] \\
    Smoothness (SE)                & 0.002  & 0.031   & 0.011 [0.003, 0.018]        & 0.389 [0.089, 0.667]& 0.182 [0.061, 0.714]& 0.207 [0.117, 0.462] \\
    Compactness (SE)               & 0.002  & 0.135   & 0.018 [0.016, 0.025]        & 0.176 [0.108, 0.251]& 0.522 [0.484, 0.657]& 0.346 [0.307, 0.466] \\
    Concavity (SE)                 & 0.000  & 0.396   & 0.021 [0.020, 0.026]        & 0.096 [0.046, 0.160]& 0.573 [0.504, 0.655]& 0.476 [0.407, 0.562] \\
    Concave Points (SE)            & 0.000  & 0.053   & 0.012 [0.009, 0.013]        & 0.191 [0.046, 0.218]& 0.632 [0.502, 0.707]& 0.441 [0.393, 0.542] \\
    Symmetry (SE)                  & 0.008  & 0.079   & 0.042 [0.011, 0.042]        & 0.360 [0.316, 0.815]& 0.857 [0.239, 0.933]& 0.497 [0.137, 0.590] \\
    Fractal Dimension (SE)         & 0.001  & 0.030   & 0.003 [0.002, 0.004]        & 0.263 [0.125, 0.317]& 0.469 [0.394, 0.572]& 0.205 [0.163, 0.338] \\
    Radius (Worst)                 & 7.93  & 36.0  & 16.8 [15.6, 17.5]     & 0.087 [0.035, 0.125]& 0.942 [0.832, 0.978]& 0.855 [0.778, 0.897] \\
    Texture (Worst)                & 12.0 & 49.5  & 24.9 [23.0, 29.3]     & 0.143 [0.064, 0.233]& 0.572 [0.498, 0.723]& 0.429 [0.374, 0.541] \\
    Perimeter (Worst)              & 50.4 & 251.2 & 105.9 [101.9, 117.2]  & 0.049 [0.019, 0.121]& 0.871 [0.829, 0.986]& 0.821 [0.786, 0.905] \\
    Area (Worst)                   & 185.2 & 4254.0 & 880.8 [739.1, 932.7] & 0.096 [0.026, 0.124]& 0.956 [0.820, 0.979]& 0.860 [0.779, 0.902] \\
    Smoothness (Worst)             & 0.071  & 0.223   & 0.136 [0.130, 0.148]        & 0.202 [0.133, 0.280]& 0.603 [0.533, 0.750]& 0.401 [0.333, 0.522] \\
    Compactness (Worst)            & 0.027  & 1.06   & 0.281 [0.201, 0.345]        & 0.175 [0.080, 0.242]& 0.760 [0.613, 0.890]& 0.585 [0.509, 0.690] \\
    Concavity (Worst)              & 0.000  & 1.25   & 0.260 [0.208, 0.349]        & 0.069 [0.017, 0.146]& 0.757 [0.675, 0.883]& 0.688 [0.633, 0.773] \\
    Concave Points (Worst)         & 0.000  & 0.291   & 0.142 [0.111, 0.149]        & 0.090 [0.034, 0.119]& 0.932 [0.826, 0.974]& 0.842 [0.777, 0.896] \\
    Symmetry (Worst)               & 0.157  & 0.664   & 0.356 [0.281, 0.360]        & 0.306 [0.172, 0.338]& 0.892 [0.558, 0.956]& 0.587 [0.351, 0.661] \\
    Fractal Dimension (Worst)      & 0.055  & 0.207   & 0.093 [0.078, 0.102]        & 0.289 [0.184, 0.334]    & 0.654 [0.492, 0.789]    & 0.365 [0.269, 0.503]    \\ \hline

    \end{tabular}%

    \end{adjustbox}
\end{table}

%% file: table_12.tex
\renewcommand{\arraystretch}{1.5}    

\begin{table}[tbp]
    \caption{Summary statistics for Heart Failure Prediction Dataset. For continuous features, the table includes minimum, maximum, and threshold values, along with conditional probabilities $P(\text{heart disease} \,|\, \text{feature} \leq \text{threshold})$ and $P(\text{heart disease} \,|\, \text{feature} > \text{threshold})$, and absolute probability difference $\Delta P$. Categorical features are maintained as categorical with probability of heart disease shown for each category. The modeled outcome is the dataset's \texttt{HeartDisease} indicator; the dataset is distributed under the name ``Heart Failure Prediction Dataset'', but the label it carries denotes presence of heart disease. Fasting blood sugar is recorded in this dataset as a binary indicator of fasting blood sugar above 120~mg/dL, and is therefore treated as categorical alongside Sex and Exercise Angina rather than being thresholded. Values in brackets denote 95\% confidence intervals estimated via bootstrap resampling ($n = 1000$).}
 
    \label{tab:heart_feature_summary_cnb}

    \begin{adjustbox}{max width=\textwidth}

    \begin{tabular}{lcccccc}
        \hline
        \multicolumn{7}{c}{\textbf{Continuous Features}}                                                                                                                                                                                                                     \\ \hline
        \multicolumn{1}{c}{\textbf{Feature}} & \textbf{Min.} & \textbf{Max.} & \textbf{$threshold$ {[}95\% CI{]}} & \textbf{$P(y{=}1 \mid x \leq t)$ {[}95\% CI{]}} & \textbf{$P(y{=}1 \mid x > t)$ {[}95\% CI{]}} & \textbf{$\Delta P$ {[}95\% CI{]}} \\ \hline
        \textbf{Oldpeak}                         & -2.60         & 6.20          & 0.600 [0.400, 1.00]               & 0.345 [0.286, 0.400]& 0.774 [0.711, 0.821]& 0.428 [0.351, 0.496] \\
        \textbf{Cholesterol}                     & 85.0        & 603.0        & 253.0 [196.0, 328.0]& 0.425 [0.327, 0.486]& 0.556 [0.482, 0.776]& 0.130 [0.109, 0.376] \\
        \textbf{MaxHR}                           & 60.0         & 202.0        & 132.0 [128.0, 157.0]         & 0.760 [0.653, 0.827]& 0.385 [0.238, 0.438]& 0.374 [0.317, 0.463] \\
        \textbf{Age}                             & 28.0         & 77.0         & 54.0 [45.0, 57.0]            & 0.418 [0.242, 0.464]& 0.696 [0.599, 0.747]& 0.277 [0.215, 0.383] \\
        \textbf{RestingBP}                       & 80.0         & 200.0        & 140.0 [118.0, 156.1]         & 0.512 [0.437, 0.567]                                     & 0.682 [0.547, 0.779]                                           & 0.170 [0.038, 0.271]       \\ \hline
        \multicolumn{7}{c}{\textbf{Categorical Features}}                                                                                                                                                                                                                   \\ \hline
        \multicolumn{1}{c}{\textbf{Feature}}     & \multicolumn{2}{c}{\textbf{Category Code}}      & \multicolumn{2}{c}{\textbf{Category Name}}                                         & \multicolumn{2}{c}{\textbf{Probability {[}95\% CI{]}}}                                          \\ \hline
        \multirow{2}{*}{\textbf{Exercise Angina}} & \multicolumn{2}{c}{1}                           & \multicolumn{2}{c}{Yes}                                                            & \multicolumn{2}{c}{0.851 [0.807, 0.892]}                                                         \\
                                                 & \multicolumn{2}{c}{0}                           & \multicolumn{2}{c}{No}                                                             & \multicolumn{2}{c}{0.349 [0.299, 0.401]}                                                         \\ \hline
        \multirow{4}{*}{\textbf{Chest Pain Type}}  & \multicolumn{2}{c}{0}                           & \multicolumn{2}{c}{ASY}                                                            & \multicolumn{2}{c}{0.792 [0.743, 0.833]}                                                         \\ 
                                                 & \multicolumn{2}{c}{3}                           & \multicolumn{2}{c}{TA}                                                             & \multicolumn{2}{c}{0.435 [0.250, 0.621]}                                                         \\
                                                 & \multicolumn{2}{c}{2}                           & \multicolumn{2}{c}{NAP}                                                            & \multicolumn{2}{c}{0.350 [0.272, 0.433]}                                                         \\
                                                 & \multicolumn{2}{c}{1}                           & \multicolumn{2}{c}{ATA}                                                            & \multicolumn{2}{c}{0.140 [0.079, 0.205]}                                                         \\ \hline
        \multirow{2}{*}{\textbf{Fasting Blood Sugar}} & \multicolumn{2}{c}{1}                      & \multicolumn{2}{c}{Yes}                                                            & \multicolumn{2}{c}{0.796 [0.732, 0.856]}                                                         \\
                                                 & \multicolumn{2}{c}{0}                           & \multicolumn{2}{c}{No}                                                             & \multicolumn{2}{c}{0.479 [0.435, 0.524]}                                                         \\ \hline
        \multirow{3}{*}{\textbf{Resting ECG}}     & \multicolumn{2}{c}{2}                           & \multicolumn{2}{c}{ST}                                                          & \multicolumn{2}{c}{0.657 [0.578, 0.745]}                                                         \\
                                                 & \multicolumn{2}{c}{0}                           & \multicolumn{2}{c}{LVH}                                                            & \multicolumn{2}{c}{0.567 [0.477, 0.651]}                                                         \\
                                                 & \multicolumn{2}{c}{1}                           & \multicolumn{2}{c}{Normal}                                                         & \multicolumn{2}{c}{0.514 [0.464, 0.570]}                                                         \\ \hline
        \multirow{2}{*}{\textbf{Sex}}            & \multicolumn{2}{c}{1}                           & \multicolumn{2}{c}{Male}                                                           & \multicolumn{2}{c}{0.630 [0.587, 0.672]}                                                         \\
                                                 & \multicolumn{2}{c}{0}                           & \multicolumn{2}{c}{Female}                                                         & \multicolumn{2}{c}{0.262 [0.184, 0.344]}                                                         \\ \hline
        \multirow{3}{*}{\textbf{ST Slope}}       & \multicolumn{2}{c}{1}                           & \multicolumn{2}{c}{Flat}                                                           & \multicolumn{2}{c}{0.828 [0.783, 0.872]}                                                         \\
                                                 & \multicolumn{2}{c}{0}                           & \multicolumn{2}{c}{Down}                                                           & \multicolumn{2}{c}{0.790 [0.660, 0.912]}                                                         \\ 
                                                 & \multicolumn{2}{c}{2}                           & \multicolumn{2}{c}{Up}                                                             & \multicolumn{2}{c}{0.198 [0.150, 0.250]}                                                         \\ \hline                                                      
        \end{tabular}

        \end{adjustbox}
\end{table}

%% file: table_13.tex
\begin{table}[H]
\caption{Leave-one-out worked example of the Bernoulli Na\"ive Bayes procedure for one Pima
Indians Diabetes sample. For each feature, the patient's measurement is compared with the
binarization threshold, and the count column $n$ gives the number of training patients of that
class falling on the \emph{same} side of the threshold as the patient. The conditional
probability is that count divided by the class size. The leave-one-out training set contains
267 diabetes patients (class prior $0.348$) and 500 non-diabetes patients (class prior
$0.652$). Thresholds can be estimated with any statistical software using the algorithm
described in Section~\ref{sec:Methods}; every remaining quantity in the table can be obtained
with a calculator.\label{tab:pima_posterior_example}}
\begin{adjustwidth}{-\extralength}{0cm}
\centering
\setlength{\tabcolsep}{4pt}
\renewcommand{\arraystretch}{1.15}
\footnotesize
\begin{tabular}{@{}l c c c | c c | c c@{}}
\toprule
 & & & & \multicolumn{2}{c|}{\textbf{Diabetes}} & \multicolumn{2}{c}{\textbf{No diabetes}} \\
\cmidrule(lr){5-6}\cmidrule(l){7-8}
\textbf{Feature} & \makecell{\textbf{Binarization}\\\textbf{threshold}} &
\makecell{\textbf{Patient}\\\textbf{value}} & \makecell{\textbf{Binarized}\\\textbf{value}} &
\textbf{\textit{n}} & $\boldsymbol{P(x\mid D)}$ &
\textbf{\textit{n}} & $\boldsymbol{P(x\mid \bar{D})}$ \\
\midrule
Pregnancies                 & 6.00  & 9.00  & 1 & 94  & 0.353 & 74  & 0.149 \\
Glucose                     & 127.0 & 171.0 & 1 & 173 & 0.647 & 109 & 0.219 \\
Blood Pressure              & 68.0  & 110.0 & 1 & 181 & 0.677 & 268 & 0.536 \\
Skin Thickness              & 31.0  & 24.0  & 0 & 165 & 0.617 & 383 & 0.765 \\
Insulin                     & 120.0 & 240.0 & 1 & 101 & 0.379 & 100 & 0.201 \\
BMI                         & 29.8  & 45.4  & 1 & 220 & 0.822 & 256 & 0.512 \\
Diabetes Pedigree Function  & 0.527 & 0.721 & 1 & 119 & 0.446 & 139 & 0.279 \\
Age                         & 28.0  & 54.0  & 1 & 196 & 0.732 & 204 & 0.408 \\
\midrule
\multicolumn{4}{@{}l}{Product of the eight conditional probabilities}
  & \multicolumn{2}{c|}{0.00970} & \multicolumn{2}{c}{0.000157} \\
\multicolumn{4}{@{}l}{Multiplied by the class prior ($0.348$ and $0.652$)}
  & \multicolumn{2}{c|}{0.00338} & \multicolumn{2}{c}{0.000102} \\
\multicolumn{4}{@{}l}{\textbf{Normalized posterior probability}}
  & \multicolumn{2}{c|}{\textbf{0.9705}} & \multicolumn{2}{c}{\textbf{0.0295}} \\
\bottomrule
\end{tabular}
\end{adjustwidth}
\end{table}

%% file: section_05_discussion.tex
\section{Discussion}\label{sec10}

The benchmark datasets used in this study, Pima Indians Diabetes, Wisconsin Breast Cancer, and Heart Failure Prediction, were selected for their broad adoption as evaluation standards in clinical machine learning, enabling direct comparison with published methods. However, the practical value of an interpretable, threshold-based framework extends well beyond these specific conditions. Consider the memory clinic setting: a nurse practitioner conducting a cognitive triage assessment needs not only a risk estimate but an accountable explanation, one that can be communicated to a worried patient or family member in real time, without reference to a black-box model or a proprietary algorithm. Similar requirements arise in community screening for type 2 diabetes in low-resource settings, in primary care cardiovascular risk communication, and in any clinical context where the decision-maker is not the data scientist who built the model. The Bernoulli Naïve Bayes framework proposed here is designed precisely for these conditions: settings where interpretability is not a desirable feature but an operational requirement, where the clinician must be able to trace, explain, and if necessary defend every classification to a patient, a colleague, or a regulatory body. Future work should evaluate the framework prospectively in such settings, including through clinician usability studies that assess whether the decision rules produced are not only formally interpretable but practically useful at the point of care.

In this context, the present study evaluated whether Bernoulli Naïve Bayes combined with supervised $\chi^2$-guided binarization can provide an interpretable option for clinical prediction without a substantial loss in discriminatory ability. The comparative analyses place the method in a competitive position relative to many conventional tabular classifiers on the three benchmark datasets. Although BNB was not uniformly superior, no statistically significant difference from a broad set of baselines was detected in a large share of the comparisons. This pattern supports the view that a threshold-based Bernoulli representation can retain useful predictive information while preserving a verifiable decision structure.

A strength of the proposed method is that the full classification pathway can be reconstructed directly from the training data. Once the thresholds have been estimated, each continuous predictor is converted into a binary rule, the relevant class-conditional probabilities can be tabulated, and the posterior probability for a new patient can be obtained with straightforward arithmetic. Table~\ref{tab:pima_posterior_example} demonstrates that the model can be presented as a complete worked example rather than as an opaque scoring mechanism. For clinical settings in which auditability and explanation are valued, this property is an important practical advantage.

The comparative results also clarify the nature of the trade-off between discrimination and interpretability. When BNB was surpassed, the gains generally came from ensemble methods, neural-network models, or selected non-linear SVM configurations. Such models can achieve stronger predictive performance, but their internal decision paths are less accessible for direct inspection at the level of an individual patient. By contrast, the proposed BNB approach expresses each predictor through an explicit threshold together with a compact set of conditional probabilities. In applications where explicit reasoning and reproducible calculation are necessary, that simplicity may justify accepting a modest reduction in raw performance.

The threshold sensitivity analysis provides further support for the proposed binarization rule. As shown in Table~\ref{tab:binarization_variants_three_datasets}, alternative thresholding strategies occasionally matched the $\chi^2$ criterion numerically, yet none achieved a statistically significant improvement in any dataset. Several alternatives were significantly worse in at least one setting. This pattern suggests that the usefulness of the method does not depend on a narrowly tuned threshold rule. Rather, the $\chi^2$ procedure appears to offer a stable and defensible way to convert continuous variables into Bernoulli features while preserving model clarity.

The calibration analysis adds a complementary perspective to the discrimination results. In clinical prediction, the quality of the probability estimates is often as important as rank-based separation, especially when probabilities may inform downstream decision thresholds. The observed improvement after beta calibration suggests that the raw Naïve Bayes probabilities can benefit from post hoc adjustment, which is consistent with the well-known tendency of conditionally independent models to produce imperfectly calibrated probabilities. From a practical standpoint, this means that the method may be used not only for classification but also for probability estimation, provided that calibration is assessed and, when needed, corrected.

The explainability analyses support the interpretability claims of the method. Figures~\ref{fig:forest_pima_features}, \ref{fig:forest_wisconsin_features}, and \ref{fig:forest_heart_features} place the selected thresholds in relation to the empirical feature distributions, while Tables~\ref{tab:pima_feature_summary}, \ref{tab:wisconsin_feature_summary}, and \ref{tab:heart_feature_summary_cnb} report outcome probabilities stratified by the selected feature thresholds or categorical levels, together with uncertainty intervals. Taken together with the patient-level posterior calculation in Table~\ref{tab:pima_posterior_example}, these components permit inspection of the model at the level of thresholds, feature-specific probabilities, and final posterior values. This level of detail is difficult to obtain from more complex classifiers without reliance on post hoc explanation methods.

The results also point to settings in which caution is warranted. Thresholding categorical predictors in the same manner as continuous variables can merge clinically meaningful categories and reduce the informational content of the original feature. A more appropriate strategy is to binarize only continuous variables while preserving categorical predictors in encoded form, as was done in the Categorical Naïve Bayes analysis for the Heart Failure dataset. As shown in the methods section, Bernoulli Naïve Bayes is the special case of Categorical Naïve Bayes in which each \emph{predictor} takes only two levels; this concerns the number of levels per feature and is independent of the number of outcome classes. Using Categorical Naïve Bayes preserves the original category structure while maintaining the simplicity of the overall approach.

Similar caution applies to the interpretation of $\Delta P$. The reported values provide an intuitive proxy for class-separation strength by describing how outcome probability changes on the two sides of a threshold, but they should not be treated as definitive measures of feature importance. Clinical predictors are often correlated, and the Naïve Bayes independence assumption does not fully represent those dependencies or their possible interactions. Consequently, the apparent contribution of one variable may change when other predictors are added or removed. Feature selection is therefore better regarded as a subset optimization problem than as a univariate ranking exercise~\cite{habibMultiobjectiveParticleSwarm2020,remeseiroReviewFeatureSelection2019,ohHybridGeneticAlgorithms2004,pudjihartonoReviewFeatureSelection2022}. Even so, the ordering of variables such as Glucose, Insulin, and Age in the Pima dataset remains clinically plausible and aligns with earlier reports~\cite{tasinDiabetesPredictionUsing2023,changPimaIndiansDiabetes2023}.

These considerations should be interpreted together with the standard limitations of Naïve Bayes. The independence assumption is unlikely to hold exactly in clinical data, the selected thresholds are sample-dependent and may vary under different resampling splits, and the present evaluation is restricted to benchmark datasets rather than external cohorts. These limitations constrain the scope of generalization. Within those bounds, the present work supports the use of statistically guided binarization as a way to make Bernoulli Naïve Bayes both transparent and practically competitive for tabular clinical prediction tasks. It should also be noted that interpretability in this study was assessed structurally, through explicit thresholds, conditional probability tables, and a worked classification example, rather than through formal evaluation with clinical users. Whether clinicians find these representations intuitive and actionable in practice remains an open question, and usability studies in real clinical settings are an important direction for future work.

%% file: section_06_conclusion_future_work.tex
\section{Conclusions}

This study evaluated a Bernoulli Naïve Bayes method based on supervised $\chi^2$-guided binarization for tabular clinical prediction. The results support its use as an interpretable reference model that combines transparent decision rules with competitive predictive performance on the evaluated datasets. The framework achieved ROC-AUC values of 0.800 for the NaN-aware Pima Indians Diabetes analysis, 0.984 for the Wisconsin Breast Cancer dataset, and 0.919 for the Heart Failure Prediction dataset. In addition to interpretability, the method remains computationally light at both training and inference, because each predictor is reduced to a single thresholded Bernoulli term and classification requires only simple probability calculations. Rather than serving as a replacement for more flexible classifiers, the method is best viewed as a simple and auditable alternative when transparency and computational efficiency are primary design goals.

The method nevertheless retains the usual limitations of Naïve Bayes. The conditional independence assumption is unlikely to hold exactly in clinical data, threshold estimates are influenced by training data variability, and mixed continuous and categorical predictors require careful handling. Exhaustive threshold evaluation over all candidate cut points may also become burdensome for larger datasets, and the absence of built-in feature weighting may limit performance when predictors differ substantially in informativeness. Strong ROC-AUC performance should therefore not be interpreted as guaranteed superiority on threshold-dependent metrics such as accuracy, precision, recall, or F1-score. The present evaluation was also limited to benchmark datasets, which constrains the scope of immediate clinical generalization.

Future work should prioritize external validation in independent cohorts, formal assessment of threshold stability, and study of feature-selection or weighting strategies that preserve interpretability. Additionally, other classifiers could benefit from similar binarization strategies, so more research is needed to explore how broadly this method can be applied.